\documentclass[A4,times,12pt]{article}

\usepackage[normalem]{ulem} 
\usepackage{times}
\usepackage{epsfig}
\usepackage{graphicx}
\usepackage{amsmath}
\usepackage{amssymb}
\DeclareMathOperator*{\argmin}{\arg \min}%
\usepackage{wrapfig}
\usepackage[dvipsnames,table]{xcolor}
\usepackage{booktabs}
\definecolor{TableBorder}{RGB}{54,76,119}
\usepackage{subfig}
\usepackage{bigints}
\usepackage{caption}
\usepackage{ulem}
\usepackage{textcomp}
\usepackage{floatpag}

\usepackage[pagebackref=true,breaklinks=true,letterpaper=true,colorlinks,bookmarks=false]{hyperref}
\newcommand\VRule[1][\arrayrulewidth]{\vrule width #1}

\title{Multiframe Motion Coupling for Video Super Resolution}

\author{Jonas Geiping\thanks{University of Siegen, \texttt{jonas.geiping@uni-siegen.de}},
Hendrik Dirks\thanks{University of M{\"u}nster, \texttt{hendrik.dirks@wwu.de}},  Daniel Cremers\thanks{Technical University of Munich, \texttt{cremers@tum.de}}, Michael Moeller\thanks{University of Siegen, \texttt{michael.moeller@uni-siegen.de}}}

\begin{document}

	\maketitle
	
	\begin{abstract}
		The idea of video super resolution is to use different view points of a single scene to enhance the overall resolution and quality. Classical energy minimization approaches first establish a correspondence of the current frame to all its neighbors in some radius and then use this temporal information for enhancement. In this paper, we propose the first variational super resolution approach that computes several super resolved frames in one batch optimization procedure by incorporating motion information between the high-resolution image frames themselves. As a consequence, the number of motion estimation problems grows linearly in the number of frames, opposed to a quadratic growth of classical methods and temporal consistency is enforced naturally. 
		
		We use infimal convolution regularization as well as an automatic parameter balancing scheme to automatically determine the reliability of the motion information and reweight the regularization locally.
		We demonstrate that our approach yields state-of-the-art results and even is competitive with machine learning approaches.
	\end{abstract}
	
	
	\section{Introduction} 
	The technique of video super resolution combines the spatial information from several low resolution frames of the same scene to produce a high resolution video.
	A classical way of solving the super resolution problem is to estimate the motion from the current frame to its neighboring frames, model the data formation process via warping, blur, and downsampling, and use a suitable regularization to suppress possible artifacts arising from the ill-posedness of the underlying problem. The final goal is to produce an enhanced, visually pleasing high resolution video in a reasonable runtime. 
	\begin{figure}[h!]
		\captionsetup[subfloat]{labelformat=empty,justification=centering,singlelinecheck=false,margin=0pt}
		\subfloat[][ Nearest, PSNR 18.63]{
			\includegraphics[trim={5.7cm 5.5cm 9.5cm 10.8cm},clip,width=0.32\textwidth]{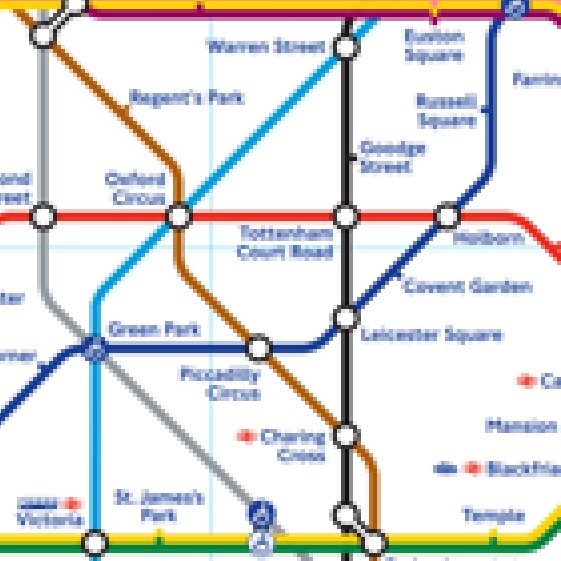}
		}
		\subfloat[][ Bicubic, PSNR 20.09]{
			\includegraphics[trim={5.7cm 5.5cm 9.5cm 10.8cm},clip,width=0.32\textwidth]{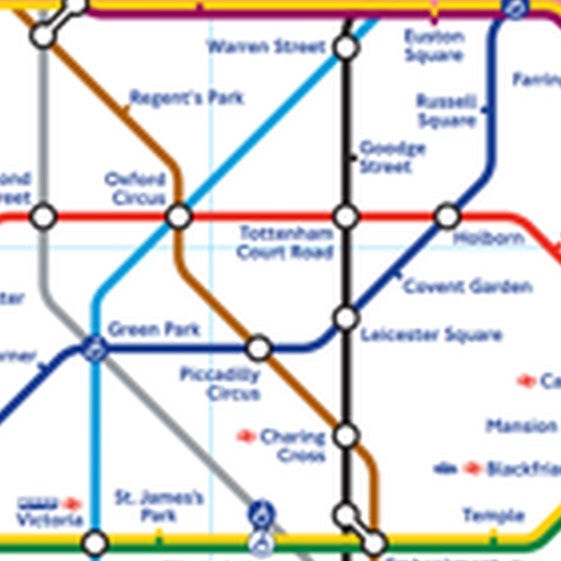}
		}
		\subfloat[][ MFSR \cite{Ma15}, PSNR 20.82]{
			\includegraphics[trim={5.7cm 5.5cm 9.5cm 10.8cm},clip,width=0.32\textwidth]{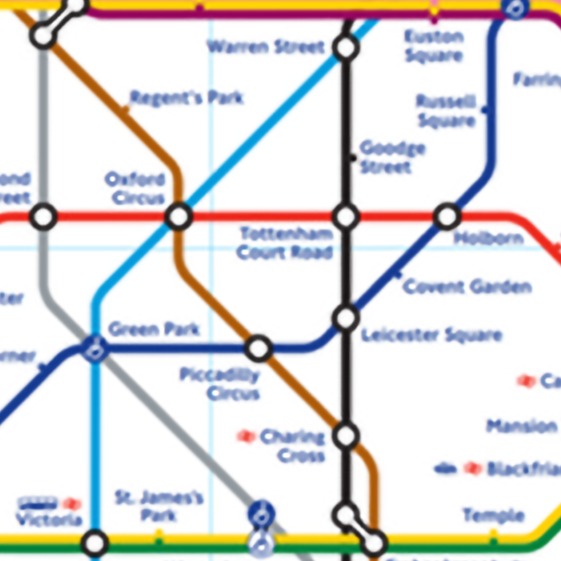}
		}\\
		
		\subfloat[][ Deep Draft \cite{liao2015video}, PSNR 21.80]{
			\includegraphics[trim={5.7cm 5.5cm 9.5cm 10.8cm},clip,width=0.32\textwidth]{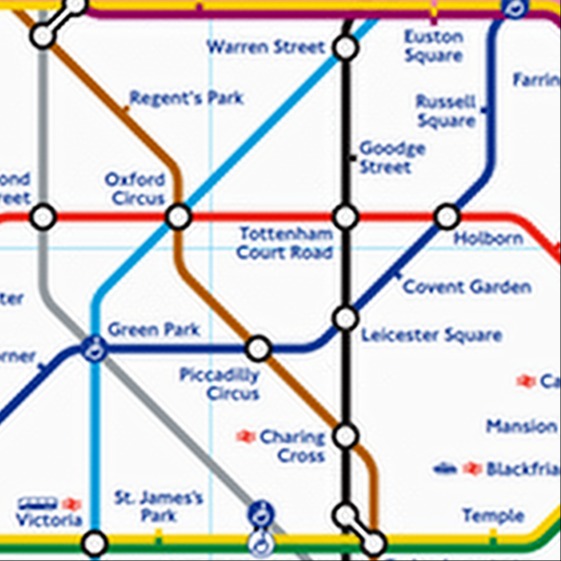}
		}
		\subfloat[][ VSRnet \cite{kappeler2016video}, PSNR 21.88]{
			\includegraphics[trim={5.7cm 5.5cm 9.5cm 10.8cm},clip,width=0.32\textwidth]{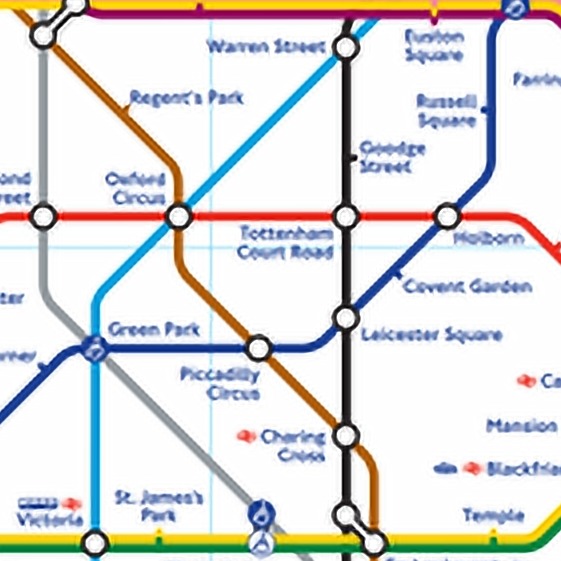}
		}
		\subfloat[][ Proposed, PSNR 23.97]{
			\includegraphics[trim={5.7cm 5.5cm 9.5cm 10.8cm},clip,width=0.32\textwidth]{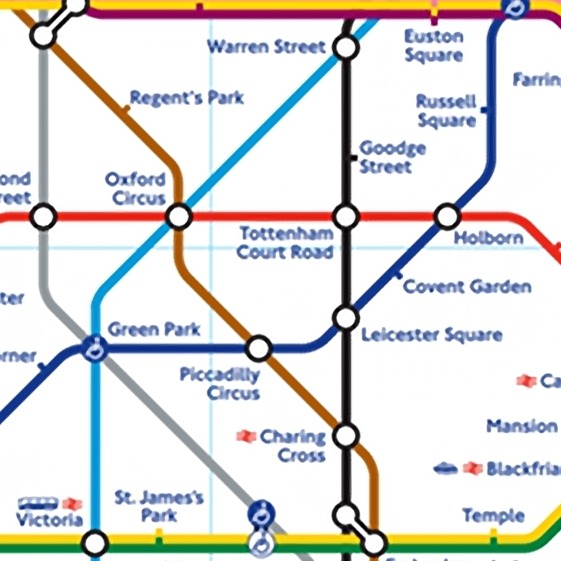}
		}\\
		\caption{\label{fig:teaser} Results for super resolving a set of 13 images of a London tube map by a factor of 4. 
			Due to the idea of jointly super resolving multiple frames, our approach behaves superior to the competing variational approach \cite{Ma15}. While approaches based on learning \cite{kappeler2016video,liao2015video} are sharp, they sometimes have difficulties resolving structures they were not trained on.}
	\end{figure}
	However the number of flow computations in this approach increases quadratically with the number of frames. Moreover, due to the strategy of super resolving each frame separately, temporal consistency cannot be enforced explicitly. Yet the latter is a key feature of a visually pleasing video: Even if a method generates a sequence of high quality high resolution frames, temporal inconsistencies will be visible as a disturbing flickering. 
	
	In addition, choosing the right strength of the regularization is a delicate issue. While a small regularization allows significant improvements in areas where the motion estimation is precise, it can lead to heavy oscillations and ringing artifacts in areas of quick motion and occlusions. A large regularization on the other hand avoids these artifacts but quickly oversmoothes the image and hence also suppresses the desirable super resolution effect. 
	
	\textbf{Contributions of this work.} We propose a method that jointly solves for all frames of the super resolved video and couples the high resolution frames directly. Such an approach tackles the drawbacks mentioned above: Because only neighboring frames are coupled \textit{explicitly}, the number of required motion estimations grows linearly with the number of frames. However by introducing this coupling on the unknown high resolution images directly, all frames are still coupled \textit{implicitly} and information is exchanged over the entire sequence. 
	
	Furthermore, we tackle the problem of choosing the right strength of spatial regularity by proposing to use the \textit{infimal convolution} between a strong spatial and a strong temporal regularization term. The latter allows our framework to automatically select the right type of regularization locally in a single convex optimization approach that can be minimized globally. 
	To make this approach robust we devise a parameter choice heuristic that allows us to process very different videos.
	
	As illustrated in Figure \ref{fig:teaser} our approach yields
	state-of-the-art results. While Figure \ref{fig:teaser} is a synthetic test consisting of
	planar motion only, we demonstrate the performance of the proposed
	approach on several real world videos in Section
	\ref{sec:results}. 

	The literature on super resolution techniques is vast and it goes beyond the scope of the paper to present a complete overview. An extensive survey of super resolution techniques published before 2012 can be found in \cite{nasrollahi2014super}. 
	We will focus on recalling some recent approaches based on energy minimization and deep learning techniques. 
	
\begin{figure}[tb]
		\centering 
		\captionsetup[subfloat]{labelformat=empty,justification=centering,singlelinecheck=false,margin=0pt}
		\subfloat[][(a) Classical coupling, e.g. \cite{unger2010convex}]{
			\includegraphics[width=0.75\textwidth]{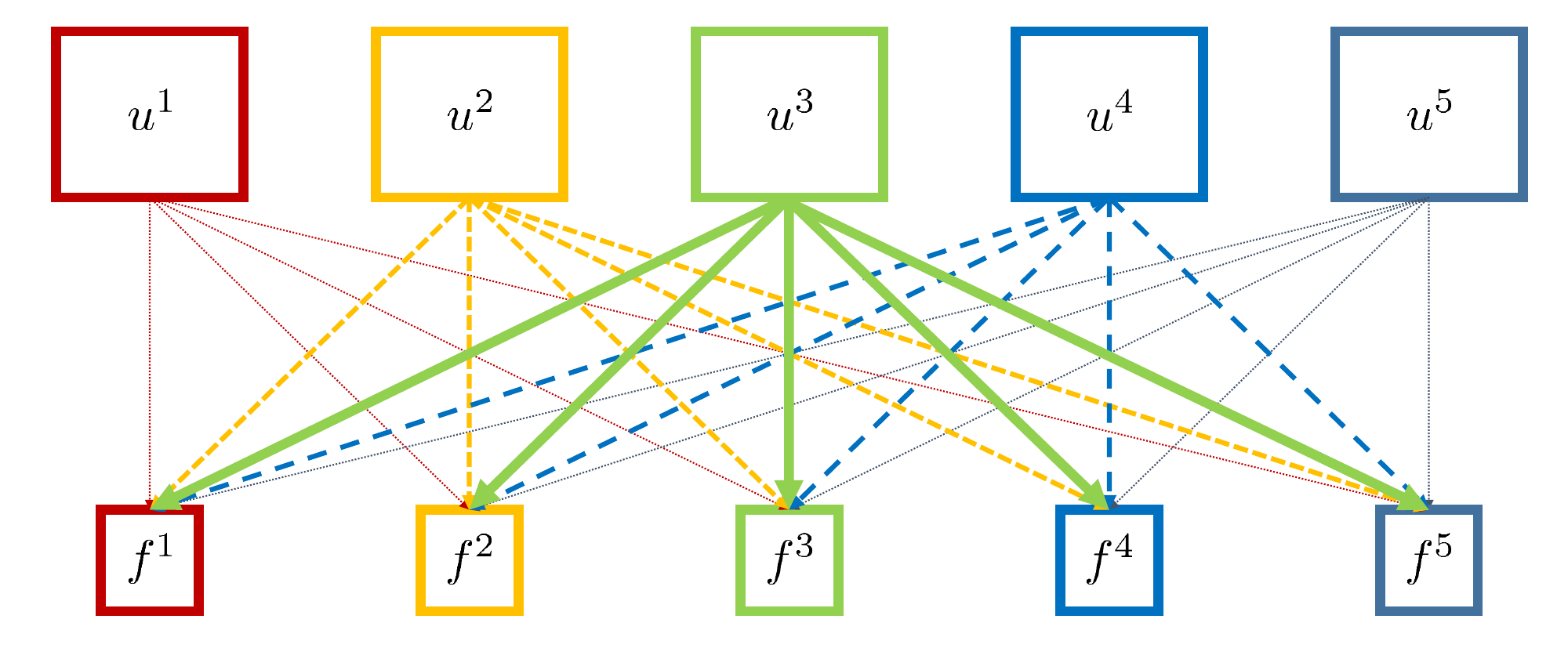}\label{fig:multiframeCouplinga}}
		\captionsetup[subfloat]{labelformat=empty,justification=centering,singlelinecheck=false,margin=0pt}
		\subfloat[][ (b) Proposed coupling in equation \eqref{eq:ourEnergy}]{\includegraphics[width=0.75\textwidth]{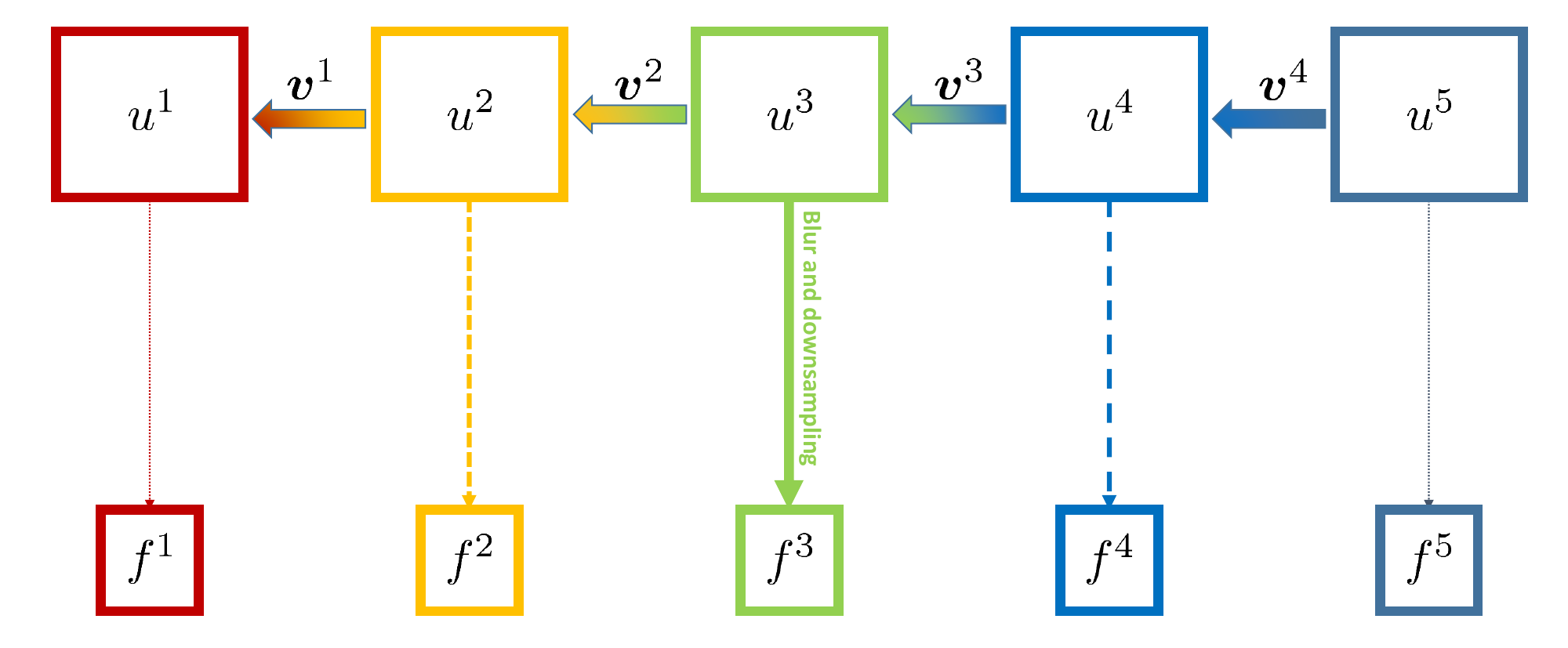}
		\label{fig:multiframeCouplingb}}
		\caption{Illustrating different kinds of temporal couplings for an examples sequence of 5 frames:  (a) shows how classical methods couple the estimated high resolution frames with all input data, (b) illustrates the proposed coupling. Each frame is merely coupled to its corresponding low resolution version.}
		\label{fig:multiframeCoupling}
	\end{figure}

	\textbf{Variational Video Reconstruction.}
	A classical variational super resolution technique was presented in \cite{unger2010convex} in which the authors propose to determine a high resolution version of the $i$-th frame via
	\begin{align}
		\label{eq:UngerEnergy}
		\min_{u^i}~ \|D(b*u^i)-f^i \|_{H^{\epsilon_d}} + \lambda \|\nabla u^i \|_{H^{\epsilon_r}} 
		+ \sum_{j \neq i}  \|D(b * W^{j,i} u^i)-f^j \|_{H^{\epsilon_d}}, 
		\end{align}
	where $\|\cdot \|_{H^{\epsilon_d}}$ denotes the Huber loss, $D$ a downsampling operator, $b$ a blur kernel, $\lambda$ a regularization parameter, and $W^{j,i}$ a warping operator that compensates the motion from the $j$-th to the $i$-th frame and is computed by an optical flow estimation in a first processing step. The temporal consistency term is based on $D(b * W^{j,i} u^i)-f^j $ and hence compares each frame to multiple low resolution frames. Figure \ref{fig:multiframeCouplinga} shows all couplings $W^{j,i}$ needed to use this approach for a sequence of frames.
	
	Mitzel et al. \cite{mitzel2009video} use a similar minimization, albeit with the $l^1$ norm instead of Huber loss. In comparison to \cite{unger2010convex} they do not compute all needed couplings but approximate them from the flows between neighboring frames, which allows for a trade-off between speed and accuracy.
	 
	Liu and Sun \cite{LiuSun13} proposed to incorporate different (global) weights $\theta_{j,i}$ for each of the temporal consistency terms in eq. \eqref{eq:UngerEnergy}, and additionally estimate the blur kernel $b$ as well as the warping operators $W^{j,i}$ by applying alternating minimization.
	In \cite{Ma15}, Ma et al. extended the work  \cite{LiuSun13} for the case of some of the low resolution frames being particularly blurry.
	Similar to \eqref{eq:UngerEnergy} the energies proposed in \cite{LiuSun13,Ma15} do not enforce regularity between the high resolution frames $u^i$ directly and require quadratically many motion estimations. Furthermore both works focus on a simplified downsampling procedure that is easier to invert than our more realistic model.
	 	
	In a recent work \cite{burger2016variational} on time continuous variational models, the authors proposed to use an optical flow penalty $\|\nabla u\cdot \boldsymbol{v}+u_t\|_1$ as a temporal regularization for joint image and motion reconstruction.
	While the optical flow term is exact in the temporally continuous setting, it would require small motions of less than one pixel to be a good approximation in a temporally discrete video.
	
	\textbf{Learning based approaches.}
	With the recent breakthroughs of deep
	learning and convolutional neural networks, researchers have promoted
	learning-based methods for super resolution \cite{kappeler2016video,liao2015video,fast16,Kim_2016_VDSR,shi2016real}.
	The focus of \cite{fast16,shi2016real} is the development of a real-time capable super resolution technique, such that we will concentrate our comparison to \cite{liao2015video}, \cite{kappeler2016video}, and \cite{Kim_2016_VDSR}, which focus on high image quality rather than computational efficiency.\\
	Note that \cite{liao2015video} and \cite{kappeler2016video} work with motion correction and require optical flow estimations. Similar to the classical variational techniques they register multiple neighboring frames to the current frame and hence also require quadratically many flow estimations. \\
	The very deep convolutional network VDSR \cite{Kim_2016_VDSR} is a conceptually different approach that does not use any temporal information, but solely relies on the training data.\\	
	
	\section{Proposed Method}
	For a sequence $f=f^1,\ldots,f^n$ of low-resolution input images we propose a multi-frame super resolution model based on motion coupling between subsequent frames. Opposed to any of the variational approaches summarized in the previous section, the energy we propose directly couples all (unknown) high resolution frames $u=u^1,\ldots,u^n$. Our method jointly computes the super resolved versions of $n$ video frames at once via the following minimization problem,
	\begin{align}
	\label{eq:ourEnergy}
	\begin{split}
	\min_{ u} ~  \sum_{i=1}^{n} &\|D(b *u^i)-f^i \|_1   
	+ \alpha \inf_{u=w+z} R_{\text{temp}}(w) + R_{\text{spat}}(z).
	\end{split}
	\end{align}
	The first term is a standard data fidelity term similar to \eqref{eq:UngerEnergy}. The key novelty of our approach is twofold and lies in the way we incorporate and utilize the motion information as well as the way we combine the temporal information with a spatial regularity assumption. The latter combines an extension of a spatio-temporal infimal convolution  technique proposed by Holler and Kunisch in \cite{HollerKunisch} with an automatic parameter balancing scheme. 
	\subsection{Spatio-Temporal Infimal Convolution and Parameter Balancing}
	\begin{figure}[tb]
		\captionsetup[subfloat]{labelformat=empty,justification=centering,singlelinecheck=false,margin=0pt}
		\subfloat[][\scriptsize (a) Original low resolution image]{
			\includegraphics[trim={4cm 0cm 15cm 10cm},clip,width=0.5\textwidth]{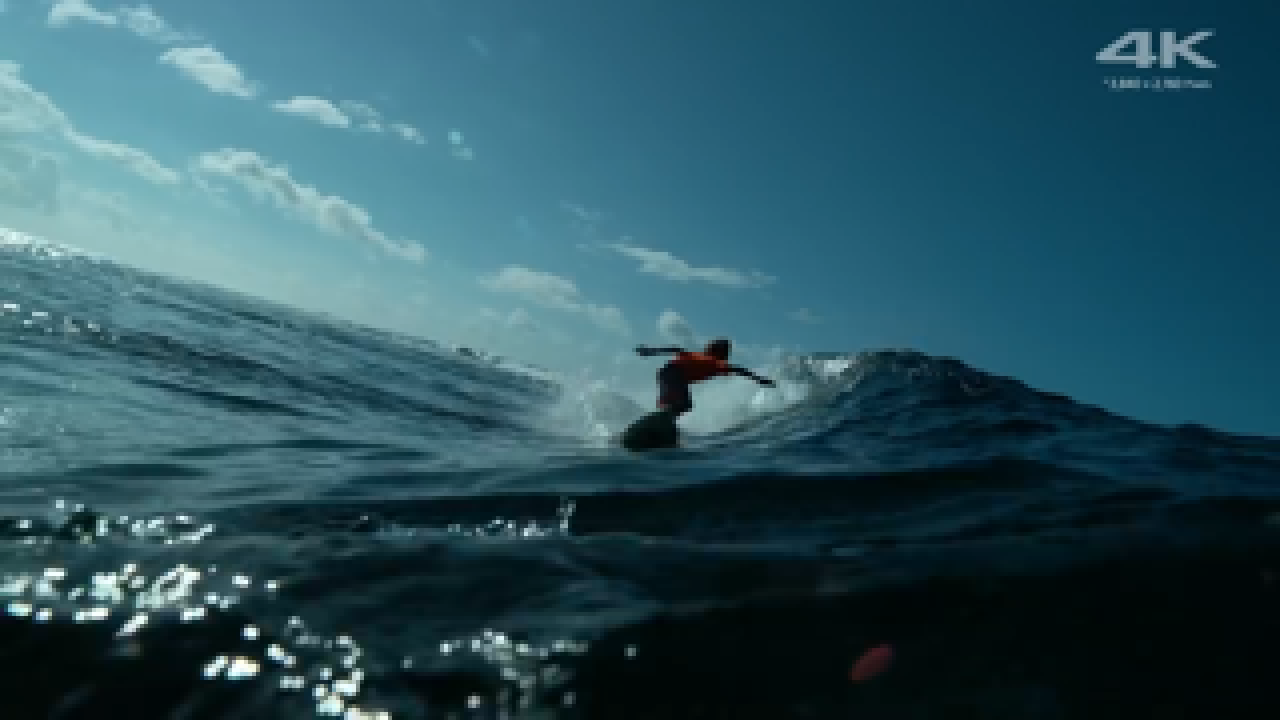}
		}
		\subfloat[][\scriptsize (b) Part $w$ with strong spatial regularization]{
			\includegraphics[trim={4cm 0cm 15cm 10cm},clip,width=0.5\textwidth]{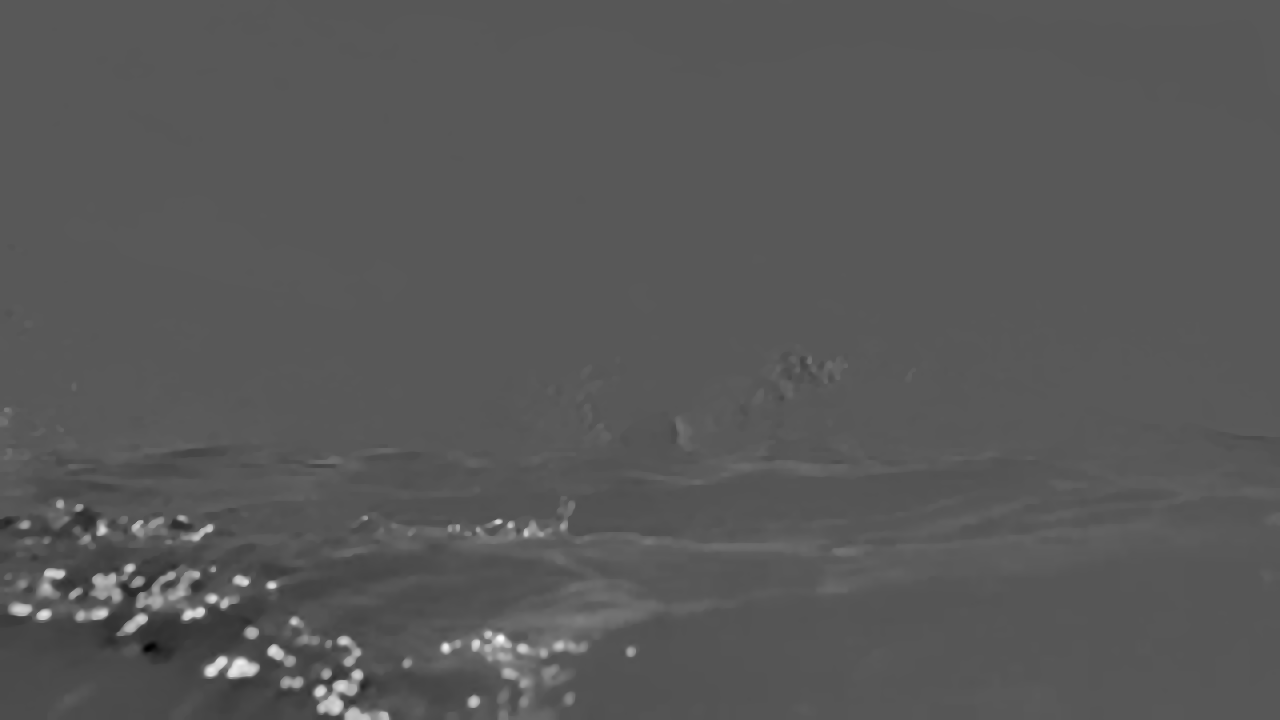}
		} \\	
		\subfloat[][\scriptsize (c) Super resolution result $u$]{
			\includegraphics[trim={4cm 0cm 15cm 10cm},clip,width=0.5\textwidth]{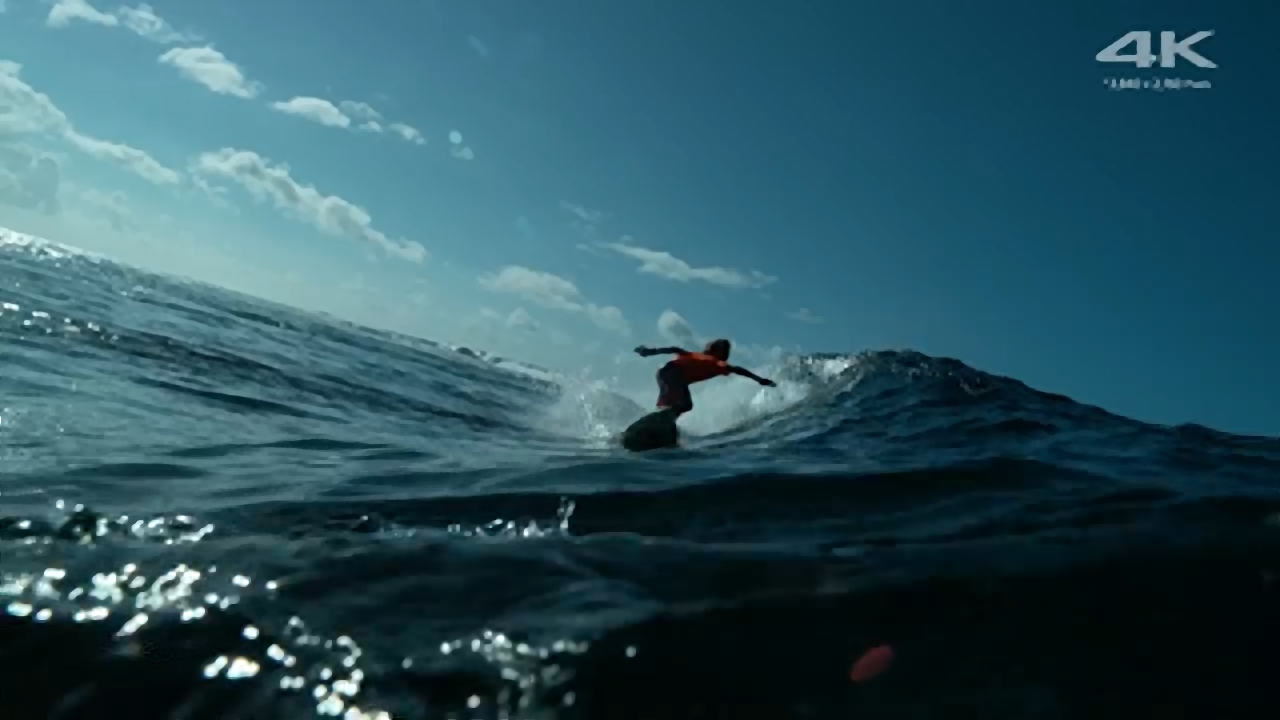}
		}
		\subfloat[][\scriptsize (d) Part $u-w$ with strong temporal regularization]{
			\includegraphics[trim={4cm 0cm 15cm 10cm},clip,width=0.5\textwidth]{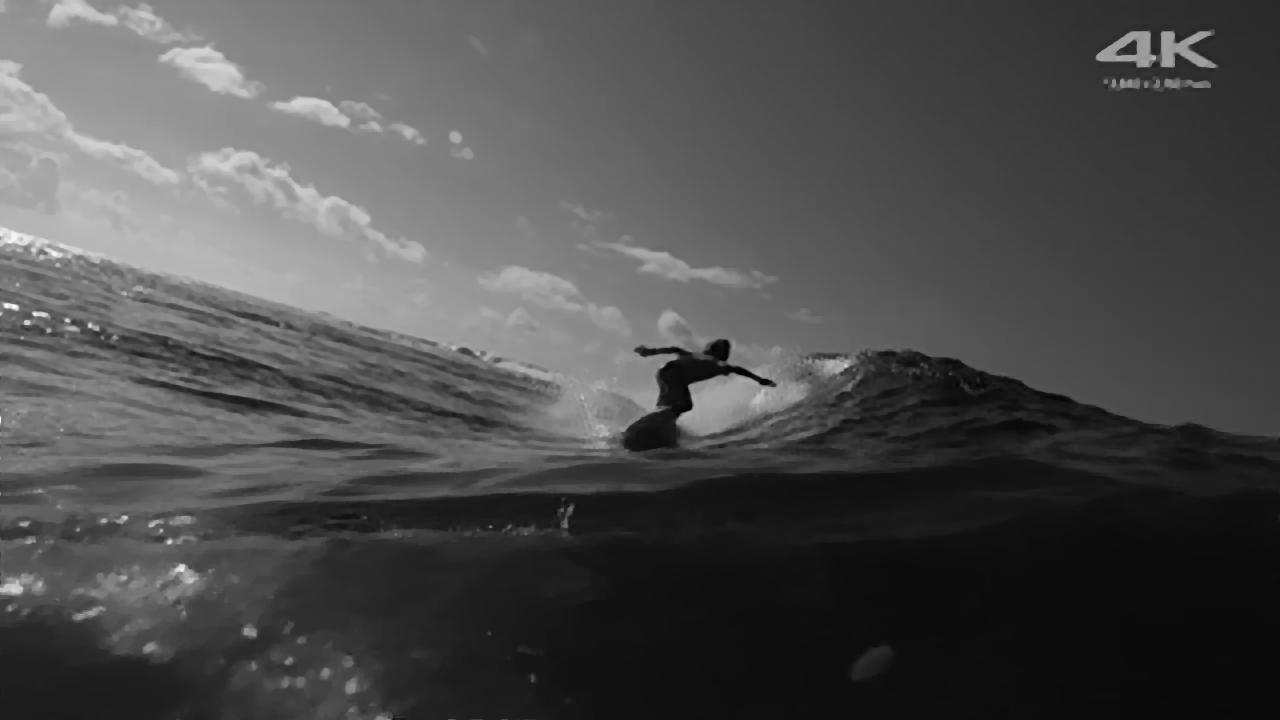}
		}\\
		\caption{Illustrating the behavior of infimal convolution regularization: The super resolution result $u$ of the low resolution input data from (a) is given in Subfigure (c). Subfigures (b) and (c) illustrate the division of $u$ into the two parts $w$ and $z=u-w$ determined by the infimal convolution regularization \eqref{eq:infConv}.
		\label{fig:infConvIllustration}}
	\end{figure}
	The second term in \eqref{eq:ourEnergy} denotes the \textit{infimal convolution}  \cite{ChambolleLions1997} between a term $R_{\text{temp}}$, which is mostly focused on introducing temporal information, and a term $R_{\text{spat}}$, which is mostly focused on enforcing spatial regularity on $u$. The infimal convolution between the two terms is defined as 
	\begin{align}
	\label{eq:infConv}
	(R_{\text{temp}} \square R_{\text{spat}})(u) := \inf_{u = w + z} R_{\text{temp}}(w) +  R_{\text{spat}}(z).
	\end{align}
	It can be understood as a convex approximation to a logical \texttt{OR} connection and allows to optimally divide the input $u$ into two parts, one of which is preferable in terms of the costs $R_{\text{temp}}$ and the other one in terms of the costs $R_{\text{spat}}$. The respective costs are defined as
	{
		\begin{align}
		R_{\text{spat}}(u) = \sum_{i=1}^{n}\left\|\sqrt{(u^i_x)^2 + (u^i_y)^2 + (\kappa  W(u^i,u^{i+1}))^2}\right\|_1,\\
		R_{\text{temp}}(u) = \sum_{i=1}^{n}\left\|\sqrt{(\kappa u^i_x)^2 + (\kappa u^i_y)^2 + (W(u^i,u^{i+1}))^2}\right\|_1,
		\end{align}
	}
	for $\kappa<1$, where the subscripts $x$ and $y$ denote the $x$- and $y$-derivatives, and $W$ denotes the photoconsistency
	{
		\begin{align}
		W(u^i,u^{i+1})(x) = 
		\frac{u^{i}(x) - u^{i+1}(x + \boldsymbol{v}^i(x))}{h}
		\end{align}}
	given a motion field $\boldsymbol{v}$, see section \ref{sec:opticalFlow}.
	The parameter $h$ encodes the scaling of time time and space dimensions  and is estimated automatically as the ratio of warp energy to gradient energy on a bicubic estimate $u_0$:
	{
		\begin{equation}
		\label{eq:hThing}
		h = \frac{ \|\mathcal{W} u_0 \|_1}{\|\partial_x u_0 \|_1 + \| \partial_y u_0 \|_1},
		\end{equation}}
	where $\mathcal{W} u_0$ denotes the vector-valued image obtained by stacking all $W(u^i, u^{i+1})$, and $\partial_x u_0, \partial_y u_0$ denote the stacked $x$- and $y$-derivatives of all frames of the sequence.
	
	Since the warp operator is multiplied with  $h^{-1}$ this provides an image-adaptive way to make sure that the spatial and temporal regularity terms are in the same order of magnitude. Note that such a term also makes sense from a physical point of view: Since $u_x$ and $u_y$ measure change in space and $W(u^i, u^{i+1})$ measures change per time, a normalization factor with units `time over space' is necessary to make these physical quantities comparable. A related discussion can be found in \cite[Section 4]{HollerKunisch}. 
	
	The idea for using the infimal convolution approach originates from \cite{HollerKunisch} in which the authors used a similar term with a time derivative instead of the operator $W$ for video denoising and decompression. 
	The infimal convolution automatically selects a regularization focusing either on space or time at each point. At points in the image where the warp energy $W(u^i,u^{i+1})$ is high, our approach automatically uses strong total variation (TV) regularization. In this sense it is a convex way of replacing the EM-based local parameter estimation from \cite{Ma15} by a joint and fully automatic regularization method with similar effects: It can handle inconsistencies in the motion field $\boldsymbol{v}$ by deciding to determine such locations by $R_{\text{spat}}$. On the other hand introducing strong spatial regularity can suppress details to be introduced by the temporal coupling. The infimal convolution approach allows favoring the optical flow information without over-regularizing those parts of the image, where the flow estimation seems to be faithful.
	
	Figure \ref{fig:infConvIllustration} demonstrates the behavior of the infimal convolution by illustrating the division of one frame into the two parts $w$ and $z=u-w$ of \eqref{eq:infConv}. Areas in which the optical flow estimation is problematic are visible in the $w$ variable and hence mostly regularized spatially. All other areas are dominated by strong temporal regularization.
	
	\subsection{Multiframe Motion Coupling}
	A key aspect of our approach is the temporal coupling of the (unknown) \textit{high resolution frames} $u$. It is based on color constancy assumptions and couples the entire sequence in a spatio-temporal manner using only linearly many flow fields $\boldsymbol{v^i}$. Figure \ref{fig:multiframeCoupling} illustrates the difference of the temporal coupling of previous energy minimization techniques and the proposed method.
	Besides only requiring linearly many flow fields, the high resolution frames are estimated jointly such that temporal consistency is enforced directly. Note that the energies \eqref{eq:UngerEnergy}, or the ones of \cite{mitzel2009video,LiuSun13,Ma15} decouple and solve for each high resolution frame separately with the temporal conformance only given by the consistency of the low resolution frames $f^i$, so that inconsistent flickering in high resolution components is not accounted for.
	
	\section{Optimization}
	The optimization is performed in a two-step procedure: We compute the optical flow on the low resolution input frames and upsample the flow to the desired resolution using bicubic interpolation. Then we solve the super resolution problem \eqref{eq:ourEnergy}.
	We experimented extensively with an alternating scheme, c.f. \cite{LiuSun13}, however the effective resolution increase through this recurring optical flow computation is marginal as we will discuss in section \ref{sec:analysis}.
	An alternative approach shown by the authors of \cite{unger2010convex} would be to compute the high resolution optical flow on a bicubic video estimate. However our experiments showed that our approach was as precise while being much more efficient. 
	\subsection{Optical Flow Estimation}
	\label{sec:opticalFlow}
	The optical flow  $\boldsymbol{v}$ on low resolution input frames $f^i$ is calculated via 
	{
		\begin{align}
		\label{subproblemV}
		\begin{split}
		\boldsymbol{v} &= \argmin_{\boldsymbol{v}}~\sum_{i=1}^{n-1}\int_\Omega \|\nabla f^{i}(x) - \nabla f^{i+1}(x + \boldsymbol{v}^{i}(x))\|_{1} \ dx  \\
		& + \int_{\Omega}|f^{i}(x) - f^{i+1}(x + \boldsymbol{v}^{i}(x))| \ dx
		+ \beta \sum_{j=1}^{2} \|\nabla \boldsymbol{v}^{i}_j\|_{H^\epsilon}.
		\end{split}
		\end{align}}
	It consists of two data terms, one that models \textit{brightness constancy} and one that models \textit{gradient constancy}, as well as a Huber penalty ($\epsilon = 0.01$) that is enforcing the regularity of the flow field. 
	Note that \eqref{subproblemV} describes a series of $n-1$ time-independent problems. To solve each of these problems we follow well-established methods \cite{sun2014quantitative,wedel2009improved,zach2007duality} and first linearize the brightness- and gradient constancy terms using a first order Taylor expansion with respect to the current estimate $\boldsymbol{\tilde{v}}^i$ of the flow field resulting in a convex energy minimization problem for each linearization. 
	We exploit the well-known iterative coarse-to-fine approach \cite{black1996robust,brox2004high} with median filtering.
	A detailed evaluation of this strategy can be found in \cite{sun2014quantitative}. We use a primal-dual algorithm with preconditioning \cite{pock2009algorithm,chambolle2011first} to solve the convex subproblems within the coarse-to-fine pyramid using the CUDA module of the FlexBox framework \cite{dirks2016flexible}.
	\subsection{Super Resolution}
	Unlike previous approaches, the super resolution problem \eqref{eq:ourEnergy} does not simplify to a series of time-independent problems, since individual frames are correlated by the flow. Consequently, the problem is solved in the whole space/time domain. First, we want to deduce that \eqref{eq:ourEnergy} can be rewritten in the form
	\begin{align}
	\begin{split}
	\argmin_{u,w} \|\mathcal{A}u-f\|_1 +\alpha \left\| \begin{pmatrix}
	\nabla w \\ \kappa \mathcal{W}w
	\end{pmatrix} \right\|_{2,1}
	 + \alpha \left\| \begin{pmatrix}
	\kappa \nabla (u-w) \\ \mathcal{W}(u-w) 
	\end{pmatrix} \right\|_{2,1} \label{subproblemUmatrix}
	\end{split},
	\end{align}
	where $u=(u^1,\ldots,u^n)$, $f=(f^1,\ldots,f^n)$, and $\mathcal{A}=\text{diag}(DB,\ldots,DB)$ denotes a linear operator, i.e. a matrix in the discrete case after vectorization of the images $u^i$, that contains the downsampling and blur operators. We use an averaging approach for the downsampling, e.g. \cite{unger2010convex} and choose the subsequent blur operator as Gaussian blur with variance dependent on the magnification factor, e.g. $\sigma^2 = 0.6$ for a factor of 4. Similarly, the gradients on $w$ and $u-w$ are block-diagonal operators consisting of the gradient operators of the single frames along the diagonal. The operator $\mathcal{W}$ is also linear and can be seen as a motion-corrected time derivative. The notation $\| \cdot \|_{2,1}$ is used to denote the sum of the $\ell^2$ norms of the vector formed by two entries from the gradient and one entry from the warping operator $\mathcal{W}$.
	
	Based on the flow fields $\boldsymbol{v}$ from the first step, we write the functions of the form $u^i(x + \boldsymbol{v}^{i-1}(x))$ as $W^{i-1} u^i$, where the $W^i$ are bicubic interpolation operators, such that $u^i(x + \boldsymbol{v}^{i-1}(x)) \approx W^{i-1}u ^i$. The final linear operator $\mathcal{W}u$ consists of $n-1$ entries of the form $u^{i} - W^{i}u ^{i+1}$ and one final block of zeros, acting as zero Neumann boundary conditions in time. 
	
	Similar to the flow problem, we used an implementation of the primal-dual algorithm in the  PROST \cite{mollenhoff2015sublabel} framework but also provide an optional binding to Flexbox \cite{dirks2016flexible}.
	Our code is publicly available on \href{https://github.com/HendrikMuenster/superResolution}{Github}\footnote{\href{https://github.com/HendrikMuenster/superResolution}{https://github.com/HendrikMuenster/superResolution}}  for the sake of reproducibility.

	\section{Numerical Results}
	\label{sec:results}
	We choose static parameters $\alpha = 0.01, \beta = 0.2$ and $ \kappa = 0.25$ across all of our different datasets and figures as we found them to yield a good and robust trade off for arbitrary video sequences for a magnification factor of 4.
	
	To be able to super resolve color videos we follow a common approach \cite{yang2010image,kappeler2016video,Kim_2016_VDSR} and transform the image sequence into a YCbCr color space and only super resolve the luminance channel Y with our variational method. The chrominance channels Cr and Cb are upsampled using bicubic interpolation. Since almost all detail information is concentrated in the luminance channel, this simplification yields almost exactly the same peak signal-to-noise ratio (PSNR) as super resolving each channel separately.
	
	To process longer videos, we use our method with frame batches in the size of a desired temporal radius and use the last computed frame from each batch as boundary value for the next batch to ensure temporal consistency.
	
	We evaluate the presented algorithm on several scenes with very different complexity and resolution. Included in our test set is one simple synthetic scene consisting of a planar motion of the London subway map (\textit{tube}), shown in Figure \ref{fig:teaser}, four common test videos \cite{LiuSun13,kappeler2016video,xiph} (\textit{calendar, city, foliage, walk}), three sequences from \cite{liao2015video,xiph} (\textit{foreman, temple, penguins}), and four sequences  from a realistic and modern UHD video sequence (\textit{sheets, wave, surfer, dog}) \cite{Sony} subsampled to 720p, that contain large non-linear motion and complex scene geometries.
	For the sake of this comparison we focused on an upsampling factor of 4, although our variational approach is able to handle arbitrary positive real upsampling factors in a straightforward fashion
	
	We evaluate nearest neighbor (NN) and bicubic interpolation (Bic), Video Enhancer \cite{VideoEnhancer2} (a commercial upsampling software), the variational approach \cite{Ma15} (MFSR), as well as the learning based techniques Deep Draft \cite{liao2015video}, VSRnet \cite{kappeler2016video}, and VDSR \cite{Kim_2016_VDSR} using code provided by the respective authors along with our proposed method and reimplementations of the variational methods \cite{unger2010convex,mitzel2009video} (with $\alpha = 0.1$). For the sake of fairness in comparison of \cite{mitzel2009video} to \cite{unger2010convex} we computed all necessary optical flows directly instead of approximating them. We consider 13 frames of the \textit{tube, city, calendar, foliage, walk} and \textit{foreman} sets and 5 frames of the larger \textit{temple, penguins, sheets, surfer, wave} and \textit{dog} sets. The PSNR and structural similarity index measure (SSIM) \cite{wang2004image} were determined for the central image of each sequence after cropping 20 pixels at each boundary. This was done so that the classical coupling methods \cite{Ma15,liao2015video,unger2010convex,mitzel2009video} are properly evaluated at the frame with maximal information in each direction for a given batch of frames.
	
	\subsection{Evaluation of proposed improvements}
	\label{sec:eval}

	We present several incremental steps in this work. To delineate the contributions of each, we will consider the average PSNR value score over our data sets in the bar plot to the right. The baseline is given by nearest neighbors interpolation with $25.61$ dB. Bicubic interpolation yields an improvement to $27.28$ dB and total variation upsampling, e.g. \cite{marquina2008image}, adds further $0.31$ dB.\\
		\setlength{\columnsep}{7pt}
		\setlength{\intextsep}{2pt}
		\begin{wrapfigure}{r}{0.5\linewidth}
			\includegraphics[width=0.5\textwidth]{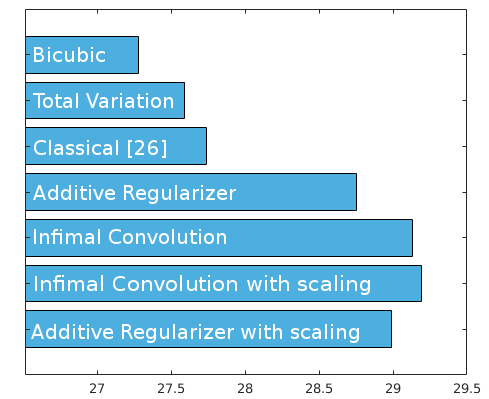}
		\end{wrapfigure}
	As a next step we consider our model of coupling frames directly, but without the infimal convolution. Instead we consider a simpler additive regularizer  first,
	\begin{equation}
	R(u) = \alpha||\mathcal{W}u||_1 + \alpha ||\nabla u||_{2,1}
	\label{eq:addcoupling}
	\end{equation}	
	using this regularizer results in an average PSNR of $28.75$ dB. It turns out that this method is already $1.01$ dB better than the classical coupling of \cite{unger2010convex}, due to failure cases in several fast-moving sets. In these cases, computing the optical flow between frames that are further apart is too error-prone, whereas the flow between neighboring frames is still reasonable to compute.
	
	 Next we consider our robustness improvements: Coupling spatial and temporal regularizers via the proposed infimal convolution \eqref{eq:infConv} increases the PSNR value by $0.38$ dB to $29.13$ dB for fixed $h=1$. Adapting the spatio-temporal scaling $h$ with the heuristic \eqref{eq:hThing} finally adds $0.06$ dB. Note this choice of $h$ can also be used directly for the additive regularizer, yielding $28.99$ dB. A memory constrained implementation of the proposed method might want to rely just on that.
	 
	 We report run times of 24 seconds per frame (\texttildelow40\% optical flow, \texttildelow60\% super resolution) for our medium sized datasets (13 frames) on a NVIDIA Titan GPU. Although these results are on a modern GPU, the flow and the super resolution problem are implemented in a general purpose framework without direct communication and with linear operators in explicit matrix notation. Further increase in speed could be obtained by porting to a specialized framework avoiding matrix representations. For comparison, our implementation of classical coupling, e.g. \cite{unger2010convex} with the same framework needs 126 (\texttildelow86\% optical flow, \texttildelow14\% super resolution) seconds per frame.
	 
	 \subsection{Choice of Forward Model}
	During our comparison to other approaches we found out that there was a significant disparity in the choice of operators for the forward model and subsequent data generation. Whereas our approach follows the works \cite{mitzel2009video,unger2010convex} and uses a bicubic downsampling process, other works \cite{LiuSun13,Ma15,liao2015video} use a Gaussian kernel followed by an asymmetric 'striding' operation, which keeps every $n$-th pixel in each direction for a downsampling factor of $n$. The Gaussian kernel in \cite{LiuSun13} is further chosen to be the theoretically optimal kernel. It turns out that this forward model is firstly easier to invert and secondly favors different strategies. Using it with our infimal convolution approach yields sharper results, significantly improving the PSNR values, e.g. the \textit{city} dataset. However the direct use of the additive regularizer, eq.\eqref{eq:addcoupling}, is the optimal choice, outperforming infimal convolution and results of \cite{LiuSun13} with up to $2.5$ dB. This is a direct consequence of the perfect match of data simulation and construction as discussed in detail in \cite[Chapter 2]{SiltanenInverseProblems}. 
	
	To have a proper evaluation, we generate data by using Matlab's bicubic image rescaling in our experiments, including color dithering and an anti-aliasing filter, followed by a clipping to obtain image values in $[0,1]$. We explicitly do not use this operator in our reconstruction, c.f. eq. \eqref{subproblemUmatrix}. Note that these shortfalls are not limited to variational methods: Neural networks equally benefit from training on exactly the same data formation process that is later used for testing.
	 
	\subsection{Comparison to other Methods}
	
	The results for all test sequences and algorithms are shown in Table \ref{ssim_table}. We structured the methods into three categories; simple interpolation based methods, variational super resolution approaches that utilize temporal information but do not require any training data, and deep learning methods. We indicate the three categories by vertical lines in the tables.
	
	Our method consistently outperforms simple interpolation techniques and also improves upon competing variational approaches, especially for complex motions like \textit{walk} or \textit{surfer}. Comparing to the learning based methods, our model based technique seems to be superior on those sequences that contain reasonable motion or a high frame rate. On sequences with particularly large motion and strong occlusions, e.g. \textit{penguins} or \textit{foreman}, the very deep convolutional neural network \cite{Kim_2016_VDSR} performs very well, possibly because it does not rely on any motion information but produces high quality results purely based on learned information. 
	
	Besides the fact that our approach remains competitive even for the aforementioned challenging data sets in terms of the PSNR values, we want to stress the importance of temporal consistency:
	Consistency of successive frames is required for a visually pleasing video perception and the lack thereof in other methods immediately yields a disturbing flickering effect.
	Demo videos showcasing this effect can be found on our \href{http://www.vsa.informatik.uni-siegen.de/en}{supplementary web page}\footnote{\href{http://www.vsa.informatik.uni-siegen.de/en/superResolution}{http://www.vsa.informatik.uni-siegen.de/en/superResolution}}, including a comparison of the consistency of our approach to the VSRnet and VDSR methods.

\begin{figure*}[hp]
		\thisfloatpagestyle{empty}
		\centering
		\captionsetup[subfloat]{labelformat=empty,justification=centering,singlelinecheck=false,margin=0pt}
		
		\subfloat[][\scriptsize \textit{calendar} dataset, ground truth zoom]{
			\includegraphics[trim={5cm 10cm 12cm 5cm},clip,height=0.15\textwidth]{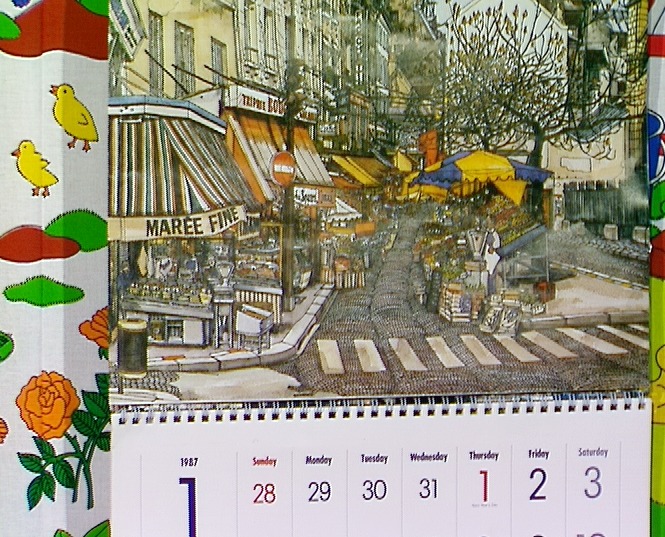}
		} 
		\subfloat[][\scriptsize \textit{walk} dataset, gound truth zoom]{
			\includegraphics[trim={12cm 9cm 4cm 3cm},clip,height=0.15\textwidth]{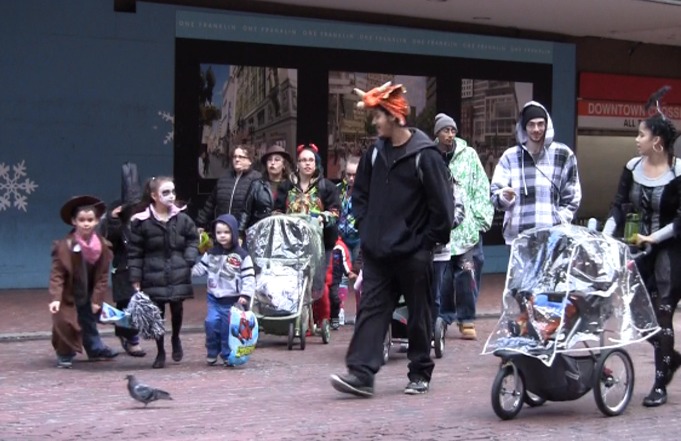}
		}
		\subfloat[][\scriptsize \textit{foreman} dataset, ground truth zoom]{
			\includegraphics[trim = {2cm 1cm 2cm 2cm}, clip,height=0.15\textwidth]{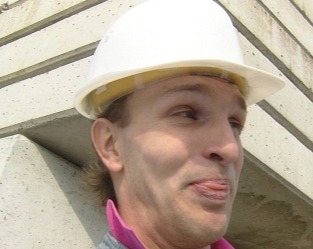}
		}
		\subfloat[][\scriptsize \textit{wave} dataset, ground truth zoom]{
			\includegraphics[trim={23cm 10cm 10cm 7cm},clip,height=0.15\textwidth]{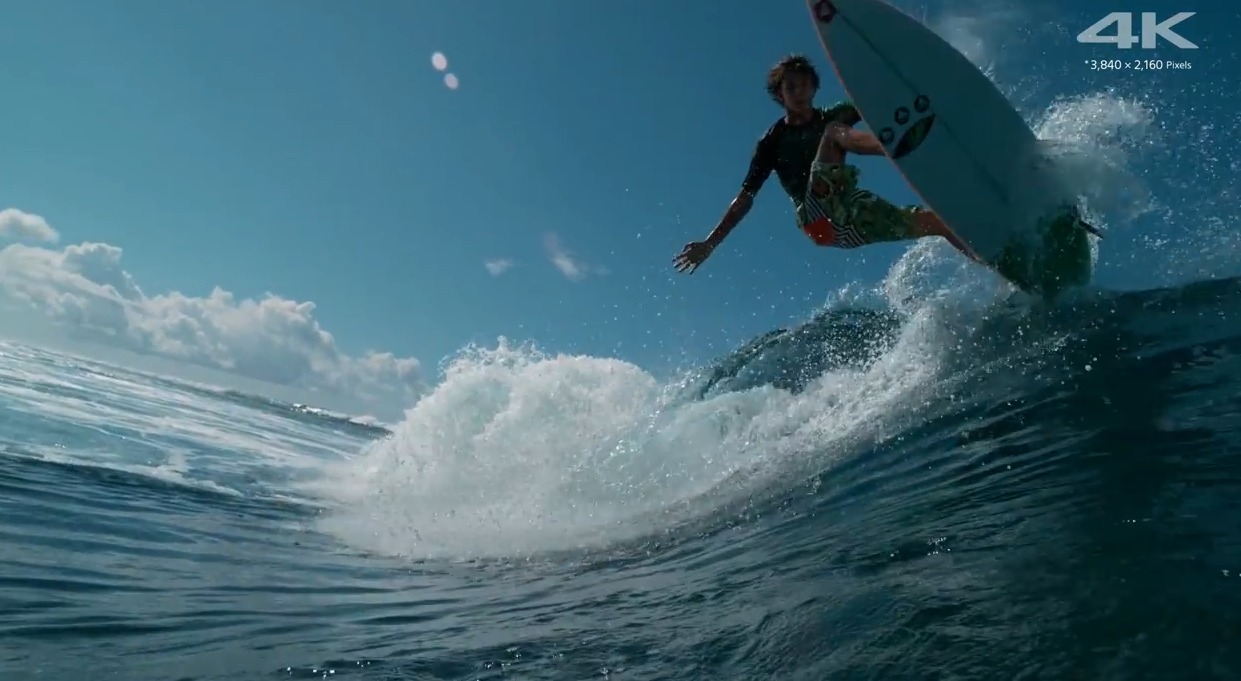}
		} \\

		\subfloat[][\scriptsize Nearest, PSNR 18.07]{
			\includegraphics[trim={5cm 10cm 12cm 5cm},clip,height=0.15\textwidth]{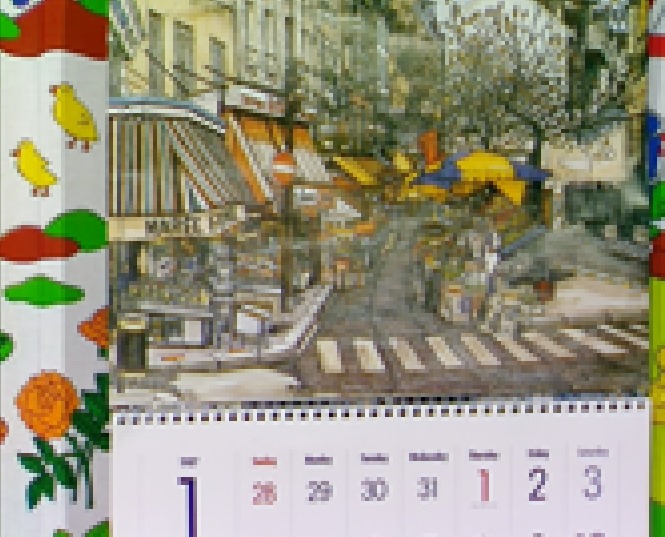}
		}
		\subfloat[][\scriptsize Nearest, PSNR 22.74]{
			\includegraphics[trim={12cm 9cm 4cm 3cm},clip,height=0.15\textwidth]{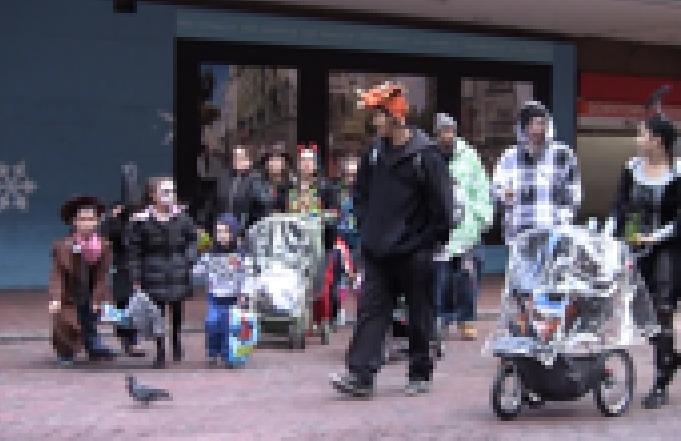}
		}
		\subfloat[][\scriptsize Nearest, PSNR 26.40]{
			\includegraphics[trim = {2cm 1cm 2cm 2cm}, clip,height=0.15\textwidth]{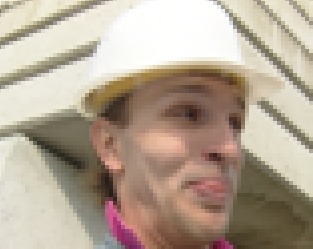}
		}
		\subfloat[][\scriptsize Nearest, PSNR 30.73]{
			\includegraphics[trim={23cm 10cm 10cm 7cm},clip,height=0.15\textwidth]{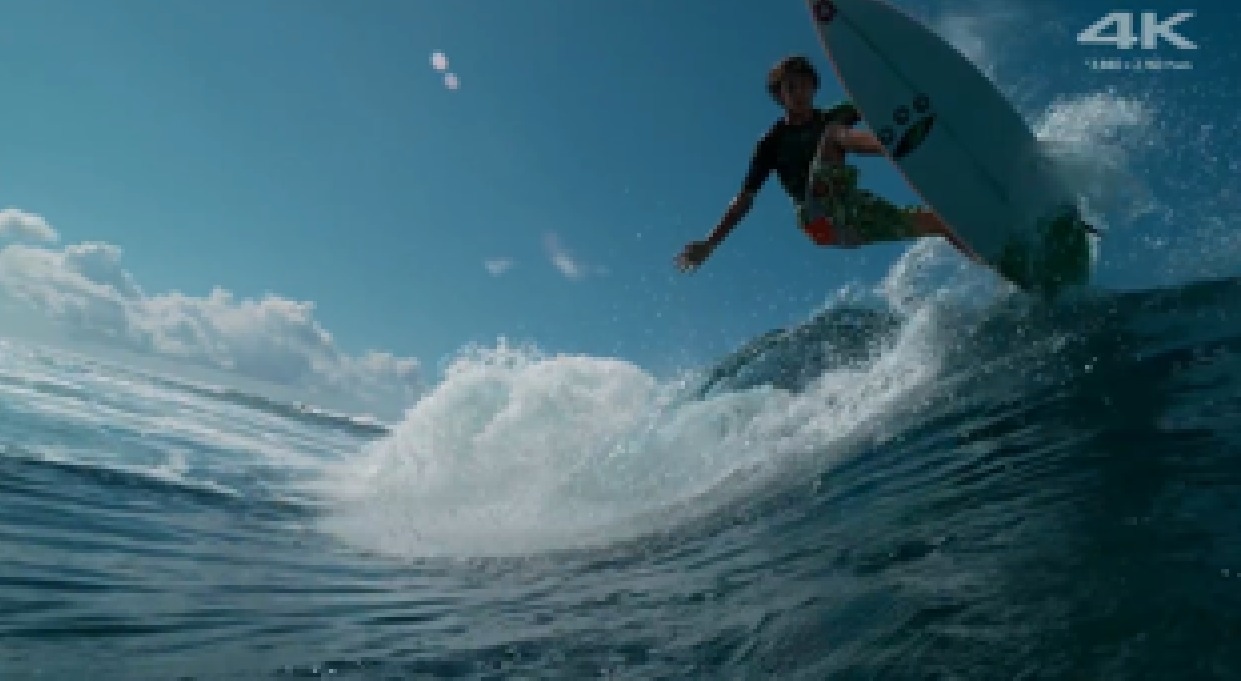}
		} \\

		\subfloat[][\scriptsize MFSR \cite{Ma15}, PSNR 19.20]{
			\includegraphics[trim={5cm 10cm 12cm 5cm},clip,height=0.15\textwidth]{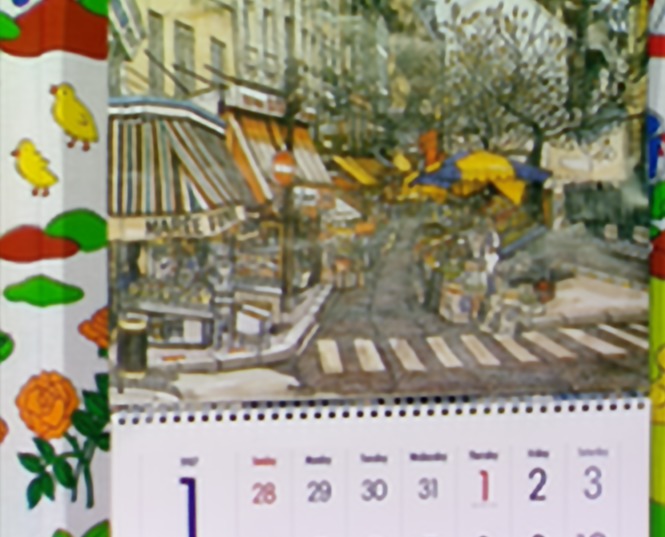}
		}
		\subfloat[][\scriptsize MFSR, PSNR 23.98]{
			\includegraphics[trim={12cm 9cm 4cm 3cm},clip,height=0.15\textwidth]{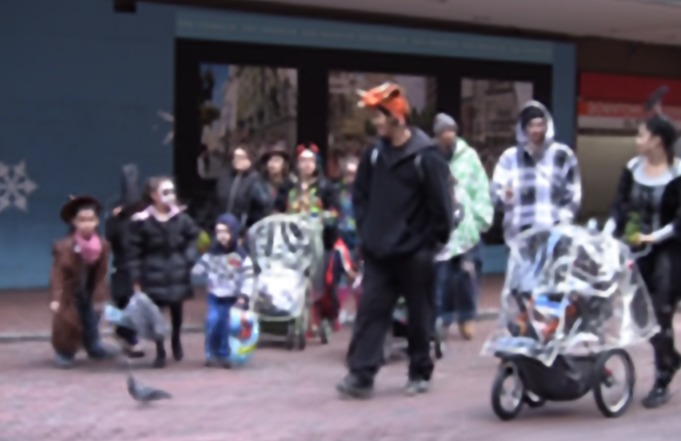}
		}
		\subfloat[][\scriptsize MFSR, PSNR 28.39]{
			\includegraphics[trim = {2cm 1cm 2cm 2cm}, clip,height=0.15\textwidth]{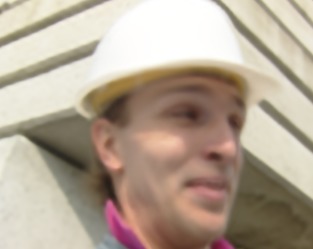}
		}
		\subfloat[][\scriptsize MFSR, PSNR 31.85]{
			\includegraphics[trim={23cm 10cm 10cm 7cm},clip,height=0.15\textwidth]{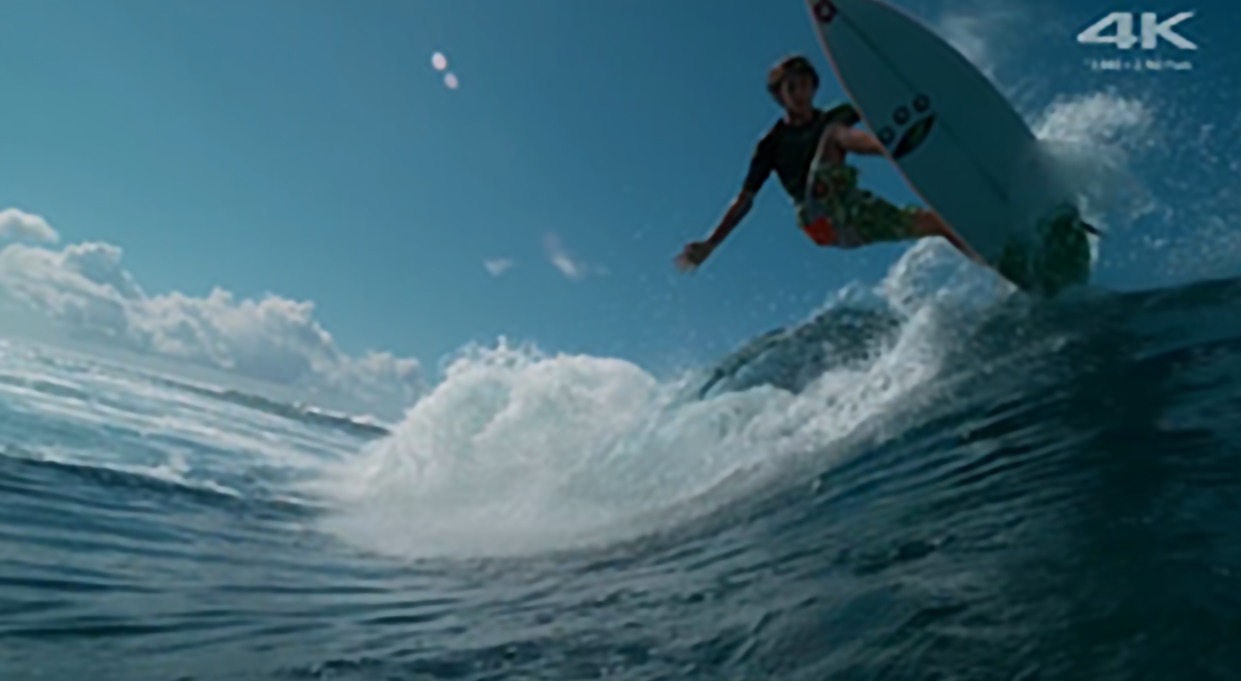}
		} \\

		\subfloat[][\scriptsize MMC (proposed method), PSNR 20.51]{
			\includegraphics[trim={5cm 10cm 12.5cm 5cm},clip,height=0.15\textwidth]{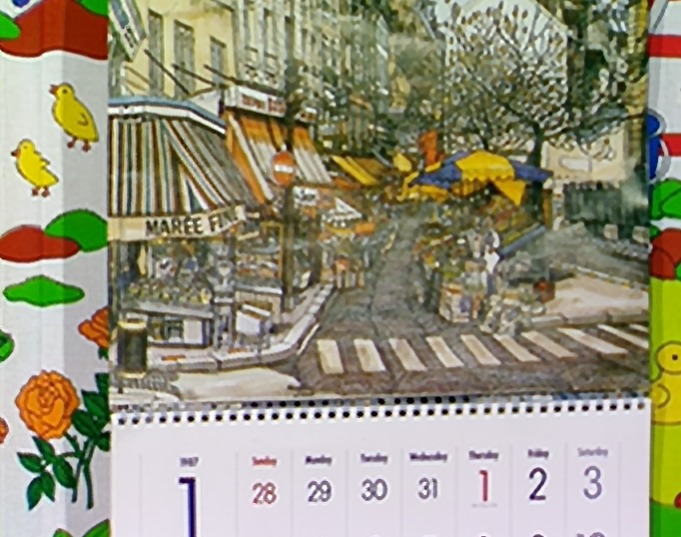}
		}
		\subfloat[][\scriptsize MMC, PSNR 26.81]{
			\includegraphics[trim={12cm 9cm 4cm 3cm},clip,height=0.15\textwidth]{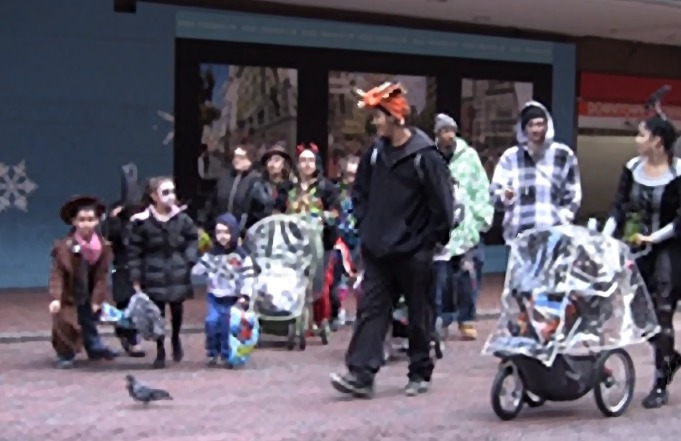}
		}
		\subfloat[][\scriptsize MMC, PSNR 31.62]{
			\includegraphics[trim = {2cm 1cm 2cm 2cm}, clip,height=0.15\textwidth]{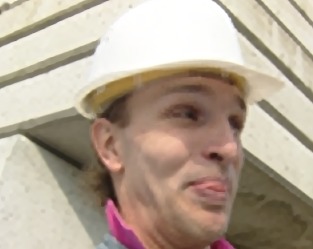}
		}
		\subfloat[][\scriptsize MMC, 33.77]{
			\includegraphics[trim={23cm 10cm 10cm 7cm},clip,height=0.15\textwidth]{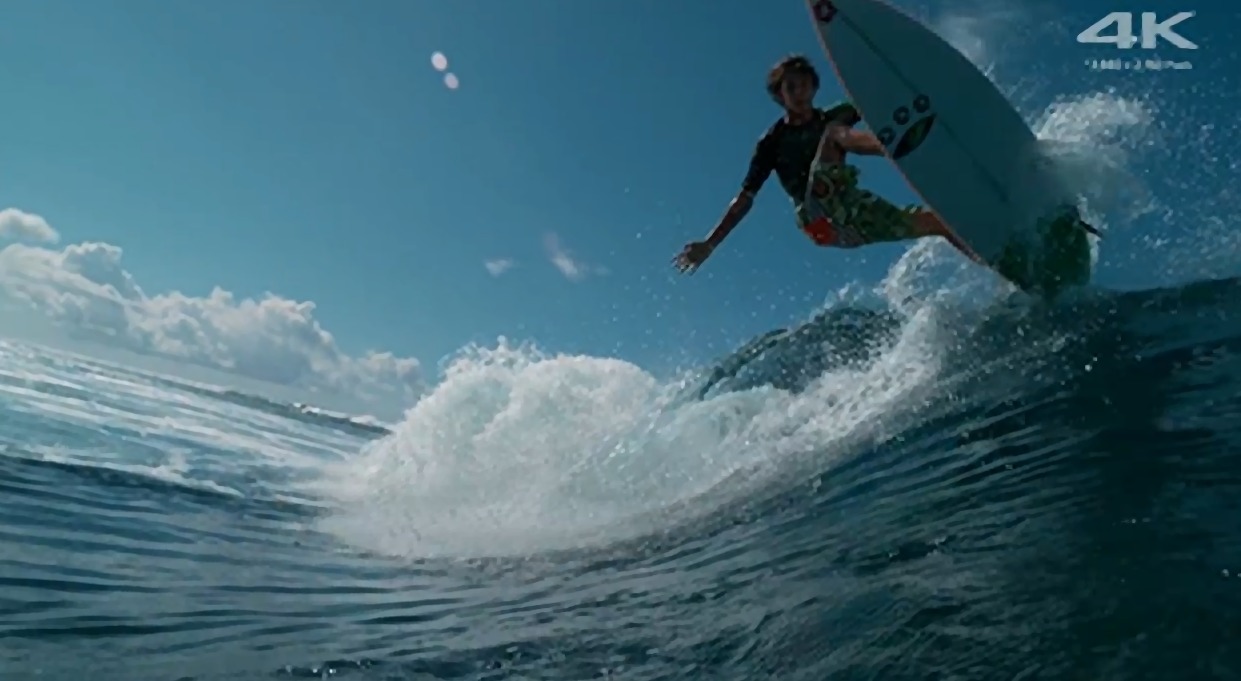}
		} \\

		\subfloat[][\scriptsize VSRnet \cite{kappeler2016video}, PSNR 19.36]{
			\includegraphics[trim={5cm 10cm 12cm 5cm},clip,height=0.15\textwidth]{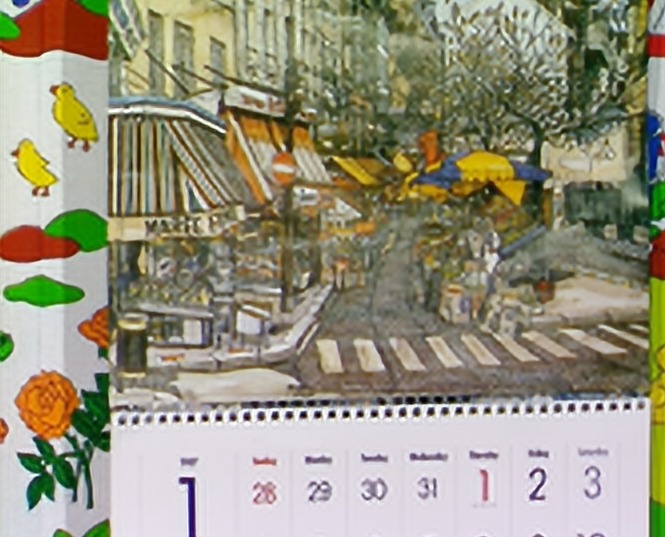}
		}
		\subfloat[][\scriptsize VSRnet, PSNR 25.95]{
			\includegraphics[trim={12cm 9cm 4cm 3cm},clip,height=0.15\textwidth]{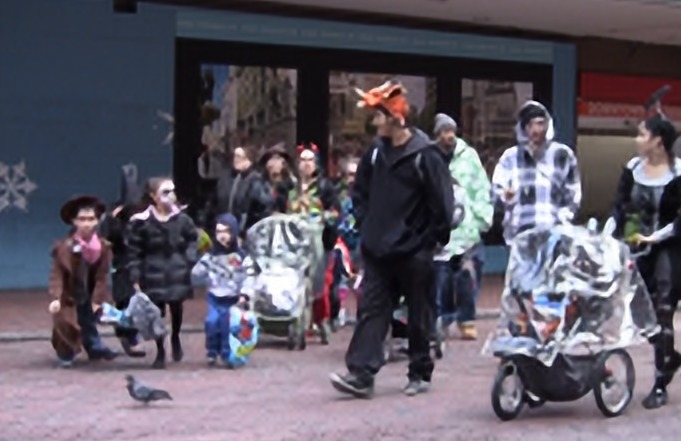}
		}
		\subfloat[][\scriptsize VSRnet, PSNR 31.02]{
			\includegraphics[trim = {2cm 1cm 2cm 2cm}, clip,height=0.15\textwidth]{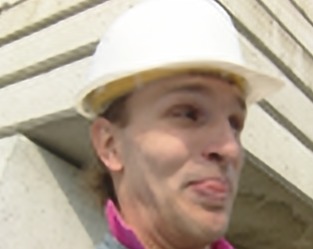}
		}
		\subfloat[][\scriptsize VSRnet, PSNR 33.03]{
			\includegraphics[trim={23cm 10cm 10cm 7cm},clip,height=0.15\textwidth]{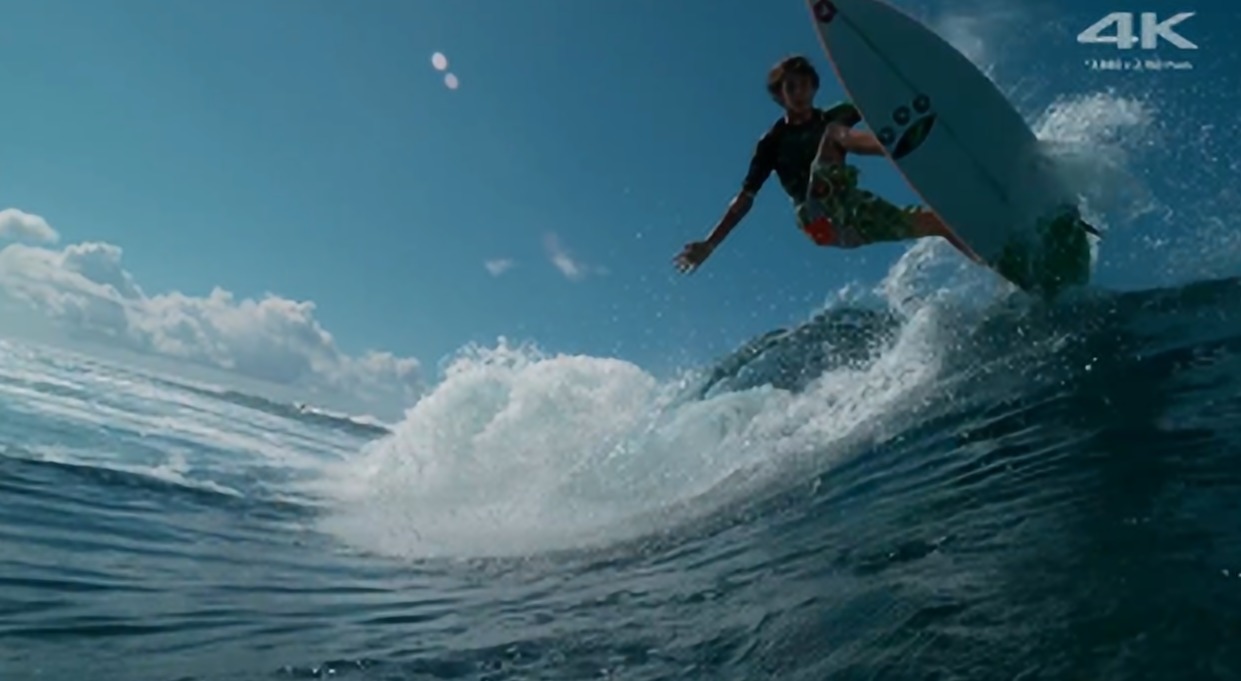}
		} \\

		\subfloat[][\scriptsize VDSR \cite{Kim_2016_VDSR}, PSNR 19.63]{
			\includegraphics[trim={5cm 10cm 12cm 5cm},clip,height=0.15\textwidth]{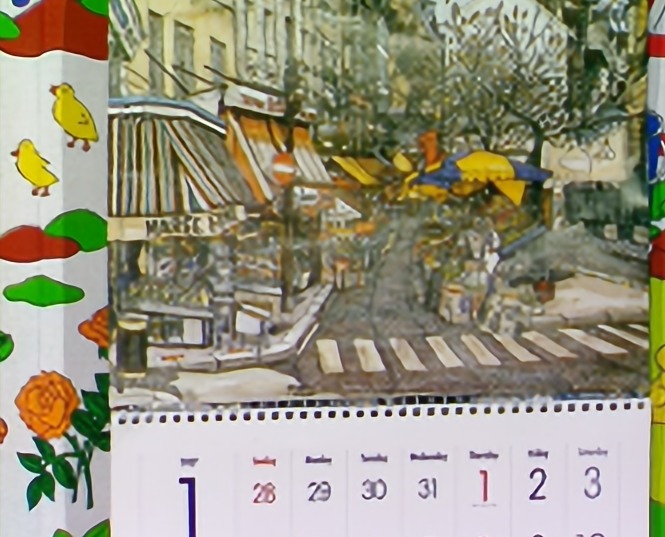}
		}
		\subfloat[][\scriptsize VDSR, PSNR 26.40]{
			\includegraphics[trim={12cm 9cm 4cm 3cm},clip,height=0.15\textwidth]{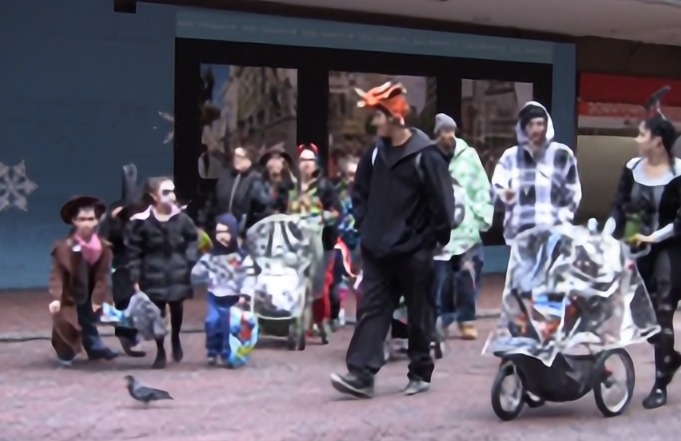}
		}
		\subfloat[][\scriptsize VDSR, PSNR 32.54]{
			\includegraphics[trim = {2cm 1cm 2cm 2cm}, clip,height=0.15\textwidth]{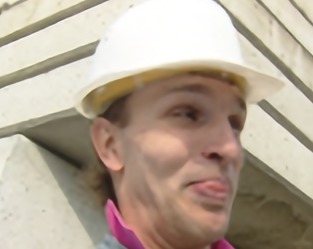}
		}
		\subfloat[][\scriptsize VDSR, PSNR 33.33]{
			\includegraphics[trim={23cm 10cm 10cm 7cm},clip,height=0.15\textwidth]{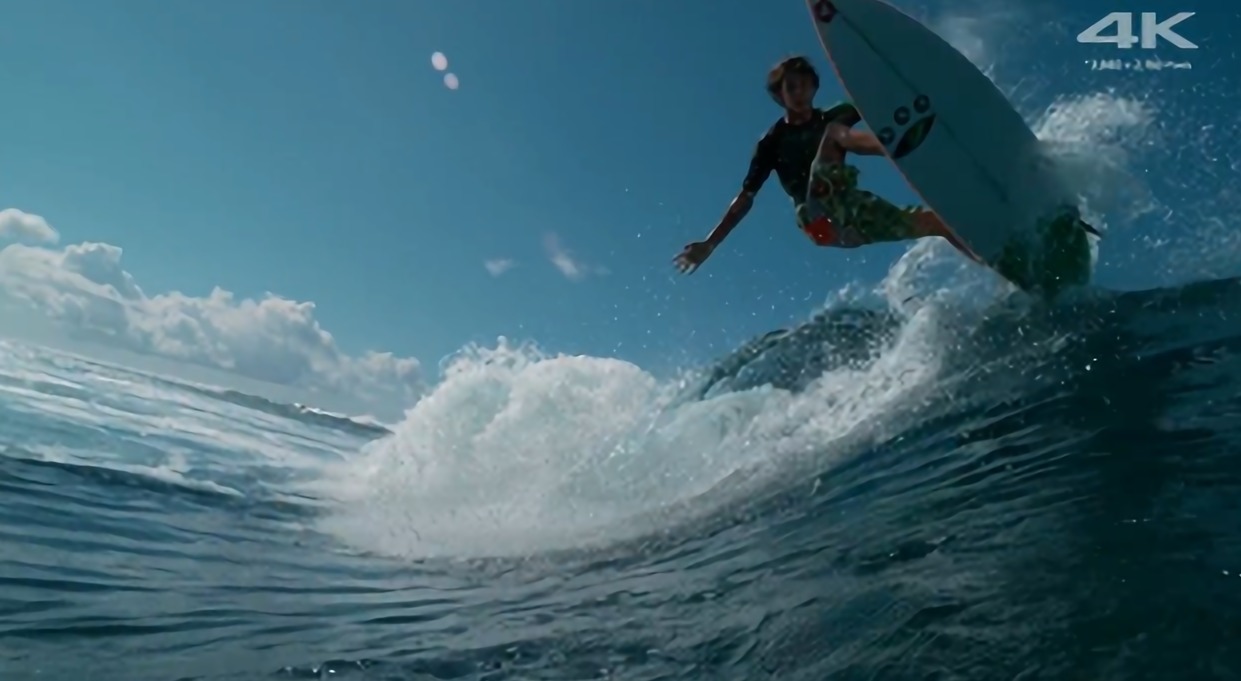}
		} \\
		
		\caption{  Super resolution by a factor of 4, zoom into datasets \textit{calendar, walk, foreman, wave}. PSNR values computed as described in Section 4. One can see the effective resolution increase of our method for the writing in \textit{calendar}, faces in \textit{walk} and wave front in \textit{wave} as well as the robustness of the approach for the challenging \textit{foreman} sequence. }
		\label{fig:compfig}
	\end{figure*}

	For a visual inspection of single frames, we present the super resolution results obtained by various methods on a selection of four data sets in Figure \ref{fig:compfig}.\\

	\begin{table}[h!]
		\centering
		\footnotesize
		\tabcolsep=0.10cm
		\rowcolors{2}{NavyBlue!15}{NavyBlue!7}
		\begin{tabular}{  
				!{\color{TableBorder}\VRule[1pt]} 
				l
				!{\color{TableBorder}\VRule[2pt]} 
				l
				l
				l
				!{\color{TableBorder}\VRule[1pt]} 
				l
				l 
				l 
				l 
				!{\color{TableBorder}\VRule[1pt]} 
				l
				l
				l
				!{\color{TableBorder}\VRule[1pt]}
			}
			\arrayrulecolor{TableBorder} \specialrule{1pt}{0pt}{0pt}
			\textbf{SSIM}& NN & Bic &  \cite{VideoEnhancer2} &  \cite{mitzel2009video}* & \cite{unger2010convex} &\cite{Ma15} & \textbf{MMC} & \cite{liao2015video} & \cite{kappeler2016video} & \cite{Kim_2016_VDSR}  \\
			\arrayrulecolor{TableBorder} \specialrule{2pt}{0pt}{0pt}
			\textit{tube}&0.800&0.846&0.898&0.943&0.937&0.877&\textbf{0.945}&0.883&0.901&0.918\\
			\textit{city }&0.596&0.634&0.702&0.760&0.745&0.653&\textbf{0.762}&0.726&0.680&0.688\\
			\textit{calendar}&0.621&0.652&0.706&\textbf{0.778}&0.764&0.686&0.772&0.738&0.705&0.726\\
			\textit{foliage}&0.760&0.797&0.809&0.859&0.857&0.809&\textbf{0.873}&0.852&0.831&0.836\\
			\textit{walk }&0.776&0.833&0.858&0.855&0.853&0.825&\textbf{0.894}&0.841&0.875&0.886\\
			\textit{foreman}&0.880&0.918&0.924&0.939&0.938&0.923&0.949&0.923&0.941&\textbf{0.953}\\
			\textit{temple}&0.835&0.874&0.893&0.910&0.909&0.878&0.924&0.820&0.916&\textbf{0.927}\\
			\textit{penguins}&0.939&0.962&0.966&0.970&0.967&0.965&0.969&0.951&0.976&\textbf{0.979}\\
			\textit{sheets}&0.948&0.971&0.978&0.978&0.978&0.972&\textbf{0.981}&0.974&0.979&0.979\\
			\textit{surfer}&0.967&0.980&0.979&0.952&0.954&0.945&0.983&0.934&0.985&\textbf{0.986}\\
			\textit{wave}&0.941&0.956&0.964&0.963&0.964&0.955&\textbf{0.971}&0.961&0.964&0.966\\
			\textit{dog}&0.955&0.971&0.974&0.971&0.972&0.970&0.975&0.970&\textbf{0.977}&\textbf{0.977}\\
			\arrayrulecolor{TableBorder} \specialrule{2pt}{0pt}{0pt}
			average &0.835&0.866&0.888&0.906&0.903&0.872&\textbf{0.917}&0.881&0.894&0.902\\
			\arrayrulecolor{TableBorder} \specialrule{2pt}{0pt}{0pt}
			\textbf{PSNR}&  &  &  &  &  &  &  &  & &  \\
			\arrayrulecolor{TableBorder} \specialrule{2pt}{0pt}{0pt}
			\textit{tube}&18.63&20.09&21.73&23.57&23.11&20.82&\textbf{23.97}&21.80&21.88&22.36\\
			\textit{city }&23.35&23.95&24.75&25.38&25.14&24.23&\textbf{25.57}&24.92&24.45&24.60\\
			\textit{calendar}&18.07&18.71&19.49&20.45&20.13&19.20&\textbf{20.51}&19.91&19.36&19.63\\
			\textit{foliage}&21.21&22.21&23.19&23.41&23.38&22.40&\textbf{24.25}&23.45&23.00&23.16\\
			\textit{walk }&22.74&24.37&25.37&24.29&24.13&23.98&\textbf{26.81}&25.00&25.95&26.40\\
			\textit{foreman}&26.40&28.66&29.31&29.51&29.13&28.39&31.62&28.95&31.02&\textbf{32.54}\\
			\textit{temple}&24.15&25.47&26.29&26.79&26.76&25.84&27.66&25.35&27.39&\textbf{27.90}\\
			\textit{penguins}&29.17&31.77&32.82&32.55&32.78&32.54&32.91&30.56&34.63&\textbf{35.00}\\
			\textit{sheets} &29.68&32.76&33.73&33.49&33.55&32.27&\textbf{34.23}&33.01&33.86&{33.85}\\
			\textit{surfer}&30.59&32.91&33.29&26.52&27.15&29.11&34.42&30.48&30.45&\textbf{34.96}\\
			\textit{wave}&30.73&31.96&32.82&32.81&32.81&31.85&\textbf{33.77}&32.43&33.03&33.33\\
			\textit{dog}&32.58&34.48&35.07&34.54&34.77&34.15&35.18&34.09&35.63&\textbf{35.71}\\
			\arrayrulecolor{TableBorder} \specialrule{2pt}{0pt}{0pt}
			average &25.61&27.28&28.16&27.77&27.74&27.07&\textbf{29.19}&27.50&28.39&29.13\\
			\arrayrulecolor{TableBorder} \specialrule{1pt}{0pt}{0pt}
		\end{tabular}
		\caption{SSIM and PSNR values (4x upsampling) from left to right: nearest neighbor, bicubic, commercial VideoEnhancer software \cite{VideoEnhancer2}, Mitzel et al. \cite{mitzel2009video} adapted for accuracy, see section \ref{sec:results}, Unger et al. \cite{unger2010convex},   Multi-Frame Super resolution \cite{Ma15}, MMC (our approach), DeepDraft ensemble learning \cite{liao2015video}, VSRnet \cite{kappeler2016video}, VDSR \cite{Kim_2016_VDSR}.}
		\label{ssim_table}
	\end{table}

	\subsection{Numerical Analysis}
	\label{sec:analysis}
	
	In light of the results of \cite{LiuSun13}, where alternating the optical flow (OF) estimation and super resolution was beneficial for a simplified and controlled data generation, we experimented with its application to our data model and more sophisticated regularization. However, as mentioned, applying the alternating procedure does not increase the video quality. The authors of \cite{Ma15} (who extend the model of \cite{LiuSun13} to include motion blur) report a similar behavior (\cite{Ma15},figure 9, $\beta = \infty$). 
	
	We investigate this further by running our approach with samples from the 
	Sintel MPI dataset \cite{sintel}, which contains ground truth OF and several levels of realism denoted by 'albedo', 'clean' and 'final', respectively. We compared PSNR values by running our method with estimated optical flow and running our method with the ground truth OF for all three realism settings, c.f. table \ref{table2}.
	\setlength{\columnsep}{7pt}
	\setlength{\intextsep}{2pt}
	\begin{wraptable}{r}{0.4\linewidth}
		\tabcolsep=0.10cm
		\rowcolors{2}{NavyBlue!15}{NavyBlue!7}
		\begin{tabular}{!{\color{TableBorder}\VRule[1pt]} 
				l  !{\color{TableBorder}\VRule[1pt]} 
				c !{\color{TableBorder}\VRule[1pt]} 
				c !{\color{TableBorder}\VRule[1pt]} }
			\arrayrulecolor{TableBorder} \specialrule{1pt}{0pt}{0pt}
			Rendering & GT Flow & our OF \\ \hline
			Albedo & 32.53 & 31.91 \\ \hline
			Clean  &  27.88 & 27.68 \\ \hline
			Final  & 33.31&  34.65\\ \hline
			\arrayrulecolor{TableBorder} \specialrule{1pt}{0pt}{0pt}
		\end{tabular}
		\caption{PSNR values computed on Sintel \cite{sintel} dataset \texttt{bandage\textunderscore1} \label{table2}}
	\end{wraptable}
	Interestingly, we do not profit from the ground truth OF on realistic data. Our super resolution warping operator $\mathcal{W}$ penalizes changes in the brightness of the current pixel to the corresponding pixels in neighboring frames, i.e. brightness constancy as does our OF. It turns out that the estimated OF yields matchings that are well suited for super resolution despite not being the physically correct ones.
	
	In light of this discussion the effectiveness of an alternating scheme is questionable. Even if the repeated OF computations converged to the GT OF, performance would not necessarily improve. The performance can only improve if the new OF would yield a refined pixel matching. \cite{Ma15} report for the case of heavy motion blur that recognizing and eliminating particularly blurry frames  can refine their matchings in an  alternating minimization. However it remains unclear how this translates into a generalized strategy, when all frames are equally low on details.	

	\section{Conclusions}
	We have proposed a variational super resolution technique based on a multiframe motion coupling of the unknown high resolution frames. The latter enforces temporal consistency of the super resolved video directly and requires only as $N-1$ optical flow estimations for $N$ frames. By combining spatial regularity and temporal information with an infimal convolution and estimating their relative weight automatically, our method adapts the strength of spatial and temporal smoothing autonomously without a change of parameters. We provided an extensive numerical comparison which demonstrates that the proposed method outperforms interpolation approaches as well as competing  variational super resolution methods, while being competitive to state-of-the-art learning approaches. For small motions or sufficiently high frame rate, our results are temporally consistent and avoid flickering effects. 
	\subsection*{Acknowledgements}
	J.G. and M.M. acknowledge the support of the German Research Foundation (DFG) via the research training group GRK 1564 Imaging New Modalities. D.C. was partially funded by the ERC Consolidator grant 3D Reloaded.
	{
		\bibliographystyle{ieee}
		\bibliography{egbib}

\begin{thebibliography}{10}\itemsep=-1pt

\bibitem{black1996robust}
M.~J. Black and P.~Anandan.
\newblock The robust estimation of multiple motions: Parametric and
  piecewise-smooth flow fields.
\newblock {\em Computer vision and image understanding}, 63(1):75--104, 1996.

\bibitem{brox2004high}
T.~Brox, A.~Bruhn, N.~Papenberg, and J.~Weickert.
\newblock High accuracy optical flow estimation based on a theory for warping.
\newblock In {\em ECCV}, pages 25--36. Springer, 2004.

\bibitem{burger2016variational}
M.~Burger, H.~Dirks, and C.-B. Sch{\"o}nlieb.
\newblock A variational model for joint motion estimation and image
  reconstruction.
\newblock {\em arXiv preprint arXiv:1607.03255}, 2016.

\bibitem{sintel}
D.~J. Butler, J.~Wulff, G.~B. Stanley, and M.~J. Black.
\newblock A naturalistic open source movie for optical flow evaluation.
\newblock In {A. Fitzgibbon et al. (Eds.)}, editor, {\em European Conf. on
  Computer Vision (ECCV)}, Part IV, LNCS 7577, pages 611--625. Springer-Verlag,
  Oct. 2012.

\bibitem{ChambolleLions1997}
A.~Chambolle and P.-L. Lions.
\newblock Image recovery via total variation minimization and related problems.
\newblock {\em Numerische Mathematik}, 76(2):167--188, 1997.

\bibitem{chambolle2011first}
A.~Chambolle and T.~Pock.
\newblock A first-order primal-dual algorithm for convex problems with
  applications to imaging.
\newblock {\em Journal of Mathematical Imaging and Vision}, 40(1):120--145,
  2011.

\bibitem{dirks2016flexible}
H.~Dirks.
\newblock A flexible primal-dual toolbox.
\newblock {\em arXiv preprint}, 2016.
\newblock http://www.flexbox.im.

\bibitem{HollerKunisch}
M.~Holler and K.~Kunisch.
\newblock On infimal convolution of tv-type functionals and applications to
  video and image reconstruction.
\newblock {\em SIAM Journal on Imaging Sciences}, 7(4):2258--2300, 2014.

\bibitem{kappeler2016video}
A.~Kappeler, S.~Yoo, Q.~Dai, and A.~K. Katsaggelos.
\newblock Video super-resolution with convolutional neural networks.
\newblock {\em IEEE Transactions on Computational Imaging}, 2(2):109--122,
  2016.

\bibitem{Kim_2016_VDSR}
J.~Kim, J.~Kwon~Lee, and K.~Mu~Lee.
\newblock Accurate image super-resolution using very deep convolutional
  networks.
\newblock In {\em The IEEE Conference on Computer Vision and Pattern
  Recognition (CVPR Oral)}, June 2016.

\bibitem{liao2015video}
R.~Liao, X.~Tao, R.~Li, Z.~Ma, and J.~Jia.
\newblock Video super-resolution via deep draft-ensemble learning.
\newblock In {\em Proceedings of the IEEE International Conference on Computer
  Vision}, pages 531--539, 2015.

\bibitem{LiuSun13}
C.~Liu and D.~Sun.
\newblock On bayesian adaptive video super resolution.
\newblock {\em IEEE transactions on pattern analysis and machine intelligence},
  36(2):346--360, 2014.

\bibitem{Ma15}
Z.~Ma, R.~Liao, X.~Tao, L.~Xu, J.~Jia, and E.~Wu.
\newblock Handling motion blur in multi-frame super-resolution.
\newblock In {\em Proceedings of the IEEE Conference on Computer Vision and
  Pattern Recognition}, pages 5224--5232, 2015.

\bibitem{marquina2008image}
A.~Marquina and S.~J. Osher.
\newblock Image super-resolution by tv-regularization and bregman iteration.
\newblock {\em Journal of Scientific Computing}, 37(3):367--382, 2008.

\bibitem{mitzel2009video}
D.~Mitzel, T.~Pock, T.~Schoenemann, and D.~Cremers.
\newblock Video super resolution using duality based {TV-L1} optical flow.
\newblock In {\em Pattern Recognition}, pages 432--441. Springer, 2009.

\bibitem{mollenhoff2015sublabel}
T.~M\"ollenhoff, E.~Laude, M.~Moeller, J.~Lellmann, and D.~Cremers.
\newblock Sublabel-accurate relaxation of nonconvex energies.
\newblock In {\em The IEEE Conference on Computer Vision and Pattern
  Recognition (CVPR)}, June 2016.
\newblock https://github.com/tum-vision/prost.

\bibitem{SiltanenInverseProblems}
J.~Mueller and S.~Siltanen.
\newblock {\em Linear and Nonlinear Inverse Problems with Practical
  Applications}.
\newblock Society for Industrial and Applied Mathematics, Philadelphia, PA,
  2012.

\bibitem{nasrollahi2014super}
K.~Nasrollahi and T.~B. Moeslund.
\newblock Super-resolution: a comprehensive survey.
\newblock {\em Machine vision and applications}, 25(6):1423--1468, 2014.

\bibitem{pock2009algorithm}
T.~Pock, D.~Cremers, H.~Bischof, and A.~Chambolle.
\newblock An algorithm for minimizing the {Mumford-Shah} functional.
\newblock In {\em Computer Vision, 2009 IEEE 12th International Conference on},
  pages 1133--1140. IEEE, 2009.

\bibitem{VideoEnhancer2}
\relax{Infognition Co. Ltd}.
\newblock Videoenhancer 2 software, version 2.1.

\bibitem{xiph}
\relax{Xiph.org, redistributable Video Test Media Collection}.
\newblock https://media.xiph.org/video/derf/.

\bibitem{shi2016real}
W.~Shi, J.~Caballero, F.~Husz{\'a}r, J.~Totz, A.~P. Aitken, R.~Bishop,
  D.~Rueckert, and Z.~Wang.
\newblock Real-time single image and video super-resolution using an efficient
  sub-pixel convolutional neural network.
\newblock In {\em Proceedings of the IEEE Conference on Computer Vision and
  Pattern Recognition}, pages 1874--1883, 2016.

\bibitem{Sony}
{\relax Sony Corporation}.
\newblock Sony 4k uhd surfing screen test demo.
\newblock CC-BY License.

\bibitem{sun2014quantitative}
D.~Sun, S.~Roth, and M.~J. Black.
\newblock A quantitative analysis of current practices in optical flow
  estimation and the principles behind them.
\newblock {\em International Journal of Computer Vision}, 106(2):115--137,
  2014.

\bibitem{unger2010convex}
M.~Unger, T.~Pock, M.~Werlberger, and H.~Bischof.
\newblock A convex approach for variational super-resolution.
\newblock In {\em Pattern Recognition}, pages 313--322. Springer, 2010.

\bibitem{wang2004image}
Z.~Wang, A.~C. Bovik, H.~R. Sheikh, and E.~P. Simoncelli.
\newblock Image quality assessment: from error visibility to structural
  similarity.
\newblock {\em Image Processing, IEEE Transactions on}, 13(4):600--612, 2004.

\bibitem{wedel2009improved}
A.~Wedel, T.~Pock, C.~Zach, H.~Bischof, and D.~Cremers.
\newblock An improved algorithm for {TV-L1} optical flow.
\newblock In {\em Statistical and Geometrical Approaches to Visual Motion
  Analysis}, pages 23--45. Springer, 2009.

\bibitem{yang2010image}
J.~Yang, J.~Wright, T.~S. Huang, and Y.~Ma.
\newblock Image super-resolution via sparse representation.
\newblock {\em IEEE transactions on image processing}, 19(11):2861--2873, 2010.

\bibitem{zach2007duality}
C.~Zach, T.~Pock, and H.~Bischof.
\newblock A duality based approach for realtime {TV-L1} optical flow.
\newblock In {\em Pattern Recognition}, pages 214--223. Springer, 2007.

\bibitem{fast16}
Z.~Zhang and V.~Sze.
\newblock Fast: Free adaptive super-resolution via transfer for compressed
  videos.
\newblock Available on ArXiv, https://arxiv.org/abs/1603.08968, 2016.

\end{thebibliography}
	}
\end{document}